\documentclass[10pt,twocolumn,letterpaper]{article}
\usepackage{color}
\usepackage[table,dvipsnames]{xcolor}
\usepackage{iccv}
\usepackage{times}
\usepackage{epsfig}
\usepackage{graphicx}
\usepackage{amsmath}
\usepackage{amssymb}
\usepackage[caption=false,font=footnotesize]{subfig}
\newcommand{\NoOne}[1]{\textcolor{red}{#1}}
\newcommand{\NoTwo}[1]{\textcolor{green}{#1}}
\newcommand{\NoThree}[1]{\textcolor{blue}{#1}}
\newcounter{RNum}
\usepackage{booktabs}
\renewcommand{\theRNum}{\arabic{RNum}}
\newcommand{\Remark}{\noindent\textit{\textbf{Remark}~\refstepcounter{RNum}\textbf{\theRNum}: }}
\usepackage[caption=false,font=footnotesize]{subfig}
\usepackage{multirow}
\usepackage{float}
\usepackage{graphics} % for pdf, bitmapped graphics files
\usepackage{epsfig} % for postscript graphics files
\usepackage{mathrsfs}
\usepackage{times} % assumes new font selection scheme installed
\usepackage{amsmath} % assumes amsmath package installed
\usepackage{amssymb}  % assumes amsmath package installed

\usepackage[pagebackref=true,breaklinks=true,colorlinks,bookmarks=false]{hyperref}
\definecolor{goodgreen}{rgb}{0,0.69,0.3137}  

\hypersetup{
	linkcolor=BrickRed
	,citecolor=Green
	,filecolor=Mulberry
	,urlcolor=NavyBlue
	,menucolor=BrickRed
	,runcolor=Mulberry
	,linkbordercolor=BrickRed
	,citebordercolor=Green
	,filebordercolor=Mulberry
	,urlbordercolor=NavyBlue
	,menubordercolor=BrickRed
	,runbordercolor=Mulberry
}
\iccvfinalcopy % *** Uncomment this line for the final submission

\def\iccvPaperID{10178} % *** Enter the ICCV Paper ID here

\ificcvfinal\fi

\begin{document}
	
	%%%%%%%%% TITLE
	\title{HiFT: Hierarchical Feature Transformer for Aerial Tracking}
	
	\author{Ziang Cao$^{\dag}$, Changhong Fu$^{\dag,}\thanks{Corresponding Author}$~, Junjie Ye$^{\dag}$, Bowen Li$^{\dag}$, and Yiming Li$^{\ddag}$\\
		$^{\dag}$Tongji University  \quad $^{\ddag}$New York University\\
		{\tt\small caoang233@gmail.com, changhongfu@tongji.edu.cn, yimingli@nyu.edu}
	}
	
	\maketitle
	\ificcvfinal\thispagestyle{empty}\fi
	
	%%%%%%%%% ABSTRACT
	\begin{abstract}
		Most existing Siamese-based tracking methods execute the classification and regression of the target object based on the similarity maps. However, they either employ a single map from the last convolutional layer which degrades the localization accuracy in complex scenarios or separately use multiple maps for decision making, introducing intractable computations for aerial mobile platforms. Thus, in this work, we propose an efficient and effective hierarchical feature transformer (HiFT) for aerial tracking. Hierarchical similarity maps generated by multi-level convolutional layers are fed into the feature transformer to achieve the interactive fusion of spatial (shallow layers) and semantics cues (deep layers). Consequently, not only the global contextual information can be raised, facilitating the target search, but also our end-to-end architecture with the transformer can efficiently learn the interdependencies among multi-level features, thereby discovering a tracking-tailored feature space with strong discriminability. Comprehensive evaluations on four aerial benchmarks have proven the effectiveness of HiFT. Real-world tests on the aerial platform have strongly validated its practicability with a real-time speed. Our code is available at \url{https://github.com/vision4robotics/HiFT}. 
	\end{abstract}
	
	%%%%%%%%% BODY TEXT
	\section{Introduction}
	
	Visual object tracking\footnote{This work targets single object tracking (SOT).}, aiming to estimate the location of object frame by frame given the initial state, has drawn considerable attention due to its prosperous applications especially for unmanned aerial vehicles (UAVs), \eg, aerial cinematography~\cite{8968163}, visual localization~\cite{9457090}, and collision warning~\cite{6907659}. Despite the impressive progress, efficient and effective aerial tracking remains a challenging task due to limited computational resources and various difficulties like fast motion, low-resolution, frequent occlusion, \etc.
	
	\begin{figure}[t]
		\centering
		\includegraphics[width=0.48\textwidth]{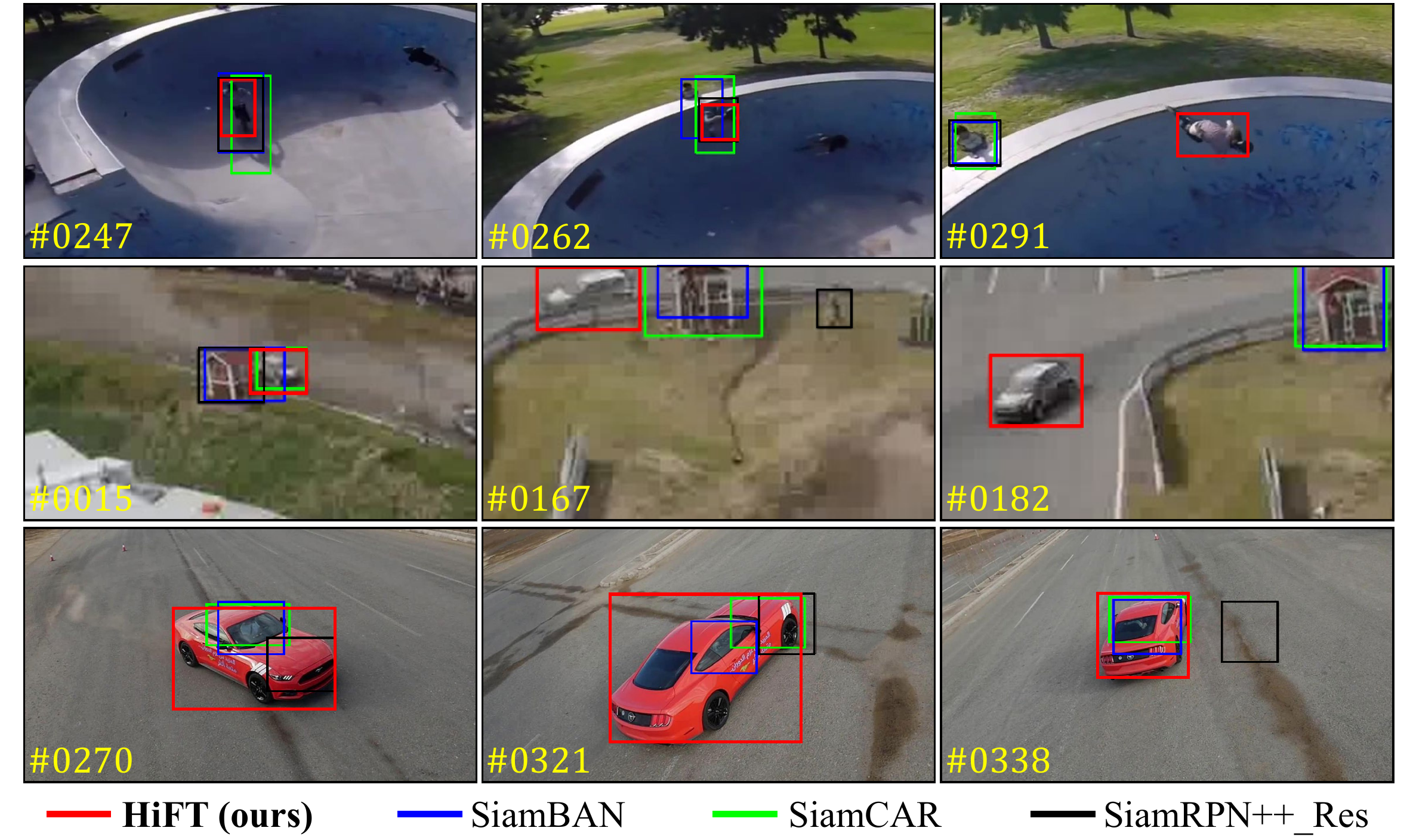}
		\caption{Qualitative comparison of the proposed HiFT with state-of-the-arts~\cite{9157720,chen2020siamese, 8954116} on three challenging sequences (\textit{BMX4}, \textit{RaceCar1} from DTB70~\cite{li2017visual}, and \textit{Car16} from UAV20L~\cite{Mueller2016ECCV}). Owing to the effective tracking-tailored feature space produced by the hierarchical feature transformer, our HiFT tracker can achieve robust performance under various challenges with a satisfactory tracking speed while other trackers lose effectiveness.}
		\label{fig:1} 
	\end{figure}
	
	In the visual tracking community, deep learning (DL)-based trackers~\cite{Wang_2019_Unsupervised,Xin2019CVPR,8100216,bertinettofully,8954116,zhu2018distractor,fu2020siamese,9477413,cao2021siamapn++} stand out on account of using the convolutional neural network (CNN) with robust representation capability. However, lightweight CNNs like AlexNet~\cite{krizhevsky2012imagenet} can hardly extract robust features which are vital for tracking performance in complex aerial scenarios. Using a larger kernel size or a deeper backbone~\cite{8954116} can alleviate the aforementioned shortcoming yet the efficiency and practicability will be sacrificed. In literature, the dilated convolution~\cite{yu2015multi} proposed to expand the receptive field and avoid the loss of resolution caused by the pooling layer. Unfortunately, it still suffers from unstable performance while handling small objects. 
	
	Recently, the transformer has demonstrated huge potential in many domains with an encoder-decoder structure~\cite{vaswani2017attention}. Inspired by the superior performance of the transformer in modeling global relationships, we try to exploit its architecture in aerial tracking to effectively fuse multi-level\footnote{We use the hierarchical feature to denote the feature maps from multiple convolutional layers.} features to achieve promising performance. Meanwhile, the loss of efficiency caused by the computations of multiple layers and the deficiency of the transformer in handling small objects (pointed out in~\cite{zhu2020deformable}) can be mitigated simultaneously. 
	
	In specific, since the target object in visual tracking can be an arbitrary object, the learned object queries in the original transformer structure hardly generalize well in visual tracking. Therefore, we adopt low-resolution features from the deeper layer to replace object queries. Meantime, we also feed the shallow layers into the transformer to discover a tracking-tailored feature space with strong discriminability by end-to-end training, which implicitly models the relationship of spatial information from high-resolution layers and semantic cues from low-resolution layers. Moreover, to further handle the insufficiency faced with low-resolution objects~\cite{zhu2020deformable}, we design a novel feature modulation layer in the transformer to fully explore the interdependencies among multi-level features. The proposed hierarchical feature transformer (HiFT) tracker has efficiently achieved robust performance under complex scenarios, as shown in Fig.~\ref{fig:1}. The main contributions of this work are as follows:
	
	\begin{itemize}
		\item  We propose a novel hierarchical feature transformer to learn relationships amongst multi-level features, thereby discovering a tracking-tailored feature space with strong discriminability for aerial tracking.
		
		\item  We design a neat feature modulation layer and classification label to further exploit the hierarchical features in Siamese networks and improve the tracking accuracy in handling the small objects.  
		
		\item Comprehensive evaluations on four authoritative aerial benchmarks have validated the promising performance of HiFT against other state-of-the-art (SOTA) trackers, even those equipped with deeper backbones.
		
		\item Real-world tests are conducted on a typical aerial platform, demonstrating the superior efficiency and effectiveness of HiFT in real-world scenarios.

	\end{itemize}
	
	%-------------------------------------------------------------------------
	\section{Related Works}
	\subsection{Visual Tracking Methods}
	After MOSSE~\cite{5539960}, a variety of achievements have been witnessed in handcrafted discriminative correlation filter (DCF)-based trackers~\cite{kiani2017learning,Li_2020_CVPR, huang2019learning,8100216}. By calculating in the Fourier domain, DCF-based trackers can achieve competitive performance with high efficiency~\cite{fu2020correlation}. Nevertheless, those trackers hardly maintain robustness under various tracking conditions due to the poor representation ability of the handcrafted feature. To improve the tracking performance, several works introducing deep learning to DCF-based methods have been released~\cite{8100216,zhang2017robust,Xin2019CVPR}. Despite the great progress, they are still faced with inferior robustness and efficiency for aerial tracking.
	%Meantime, there are amount of researches about the combination of DL-based and DCF-based methods~\cite{8100216,zhang2017robust,Xin2019CVPR}. ATOM~\cite{8953466}, DiMP~\cite{9010649}, and PrDiMP~\cite{9157124} also raise the performance to a new level. 
	%\textbf{Siamese-based trackers}:
	
	Another outstanding branch in the SOT community is the Siamese-based methods~\cite{bertinettofully,guo2017learning,8579033,zhu2018distractor,8954116}, which benefit from massive offline training data and end-to-end learning strategy. As one of the pioneering works, SiameseFC~\cite{bertinettofully} exposed the advantage of the Siamese framework, formulating the tracking task as the similarity matching process of template and search patches. Based on SiameseFC, DSiam~\cite{guo2017learning} was proposed to effectively handle the object appearance variation and background interference. Inspired by region proposal network (RPN)~\cite{girshick2015fast}, SiamRPN~\cite{8579033} considered tracking as two subtasks, applying the classification and regression branches respectively. DaSiamRPN~\cite{zhu2018distractor} introduced a novel distractor-aware module and an effective sampling strategy, further promoting its robustness. More recently, the potential of adopting very deep networks as the backbone is extensively tapped~\cite{8954116}, while the efficiency is sacrificed largely. Obviously, RPN-based trackers~\cite{8579033,zhu2018distractor,8954116} provide an effective tracking strategy. However, the hyper-parameters associated with anchors significantly decrease the generalization of trackers. In order to eliminate such a drawback, the anchor-free method is proposed~\cite{9157720,chen2020siamese}.
	\begin{figure*}[t]
		\centering
		% 调整比例，添加图片的相对位置
		%\includegraphics[scale=0.5]{images/1.pdf}
		\includegraphics[width=1\textwidth]{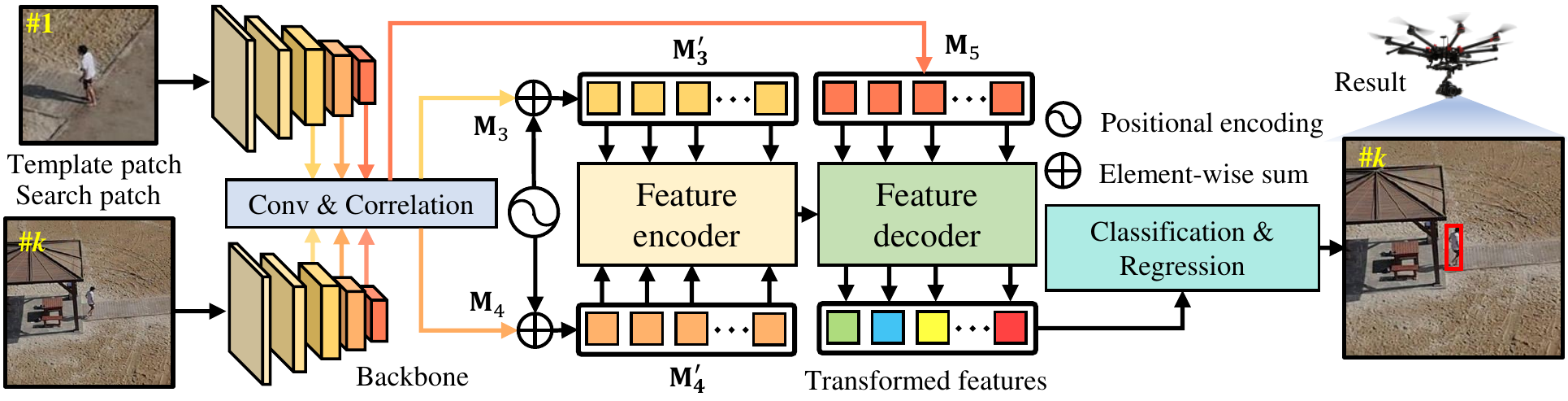}
		%\scalebox{0.5}[0.5]{\includegraphics[trim={0 75 0 55},clip]{images/1.pdf}}
		% 写标题
		\caption{Overview of the HiFT tracker. The modules from the left to right are feature extraction network, hierarchical feature transformer, and classification \& regression network. Three arrows with different colors represent the workflow of features from different layers respectively. Note that only the input of the encoder is combined with position encoding. Best viewed in color. (Image frames are from UAV20L~\cite{Mueller2016ECCV}.)
		}
		
		\label{fig:main} % 写 \label 跟在 \caption 后面，之后使用 \ref{}引用
		\vspace{-10pt}
	\end{figure*}
	
	In Siamese-based trackers, robust features make a vital influence on tracking performance. However, the trackers~\cite{bertinettofully,8579033,zhu2018distractor,fu2020siamese} with lightweight backbone like AlexNet~\cite{krizhevsky2012imagenet} suffer from the lack of global context while the trackers~\cite{8954116,chen2020siamese,9157720} utilizing deep CNN like ResNet~\cite{He2016res} are far from real-time requirements onboard UAV. Albeit several works proposed to explore multi-level features in visual tracking~\cite{8954116,8954057}, they introduce cumbersome computation inevitably which is unaffordable for mobile platforms. Differently, this work proposes a brand-new lightweight hierarchical feature transformer (HiFT) for effective and efficient multi-level feature fusion, achieving robust aerial tracking efficiently.

	\subsection{Transformer in Computer Vision}
	Vaswani \textit{et al.}~\cite{vaswani2017attention} firstly proposed the transformers for machine translation based on the attention mechanism. Benefiting from its high representation ability, the transformer structure is expanded to the domain of computer vision such as video captioning~\cite{8579009}, image enhancement~\cite{yang2020learning}, and pose estimation~\cite{huang2020hand}. After DETR~\cite{carion2020end} initiates the research of transformer in object detection, deformable DETR~\cite{zhu2020deformable} proposed the deformable attention module for efficiently convergence, providing inspirations about the combination of transformer and CNN. Some studies attempted to introduce the transformer to multi-object tracking and achieved promising performance~\cite{meinhardt2021trackformer}, while the study of transformer in SOT is still blocked so far.
	%, image generation~\cite{parmar2018image},
	%, classification task~\cite{dosovitskiy2020image}
	
	Although the attention mechanism in the transformer shows good performance in extensive visual tasks, its superiority struggles to be extended to SOT, since predefined (or learned) object queries hardly maintain effectiveness when facing an arbitrary object. Moreover, the transformer hardly deals with the low-resolution object which is frequently encountered in aerial tracking. In this work, instead of redesigning object queries and related structures, we propose a hierarchical feature transformer to constructing a novel as well as robust tracking-tailored feature space. By virtue of the introduction of global context and interdependencies among multi-level features, the discriminability in the feature space is significantly raised to improve the tracking performance. Meanwhile, HiFT possesses a lightweight encoder-decoder structure which is desirable for mobile platforms.

	\section{Proposed Method}\label{pro}
	The workflow of HiFT is presented in Fig.~\ref{fig:main}. It can be divided into three submodules, feature extraction network, hierarchical feature transformer, and classification \& regression network. Note that we utilize features from the last three layers to build the hierarchical feature transformation in this paper. 
	
	%feature maps $\mathbf{M}_3$ and $\mathbf{M}_4$ are obtained. Reshaped and supplemented with a learnable positional encoding, the two high-resolution feature maps are input into the feature-cross encoder. With reshaped low-resolution $\mathbf{M}_5$, transformed features are finally achieved by feature decoder for classification and regression network to estimate the result box.
	%
	%

	\subsection{Feature Extraction Network}
	Deep CNNs, \textit{e.g.}, ResNet~\cite{He2016res}, MobileNet~\cite{sandler2018mobilenetv2}, and GoogLeNet~\cite{szegedy2015going}, have demonstrated their surprising capability, serving as popular feature extraction backbones in Siamese frameworks~\cite{8954116}. However, the heavy computation brought by the deep structure hardly be afforded by the aerial platform. To this concern, HiFT adopts a lightweight backbone, \textit{i.e.}, AlexNet~\cite{krizhevsky2012imagenet}, which serves in both template and search branches. For clarity, the template/search images are respectively denoted by $\mathbf{Z}$ and $\mathbf{X}$. $\phi_{k}(\mathbf{X})$ represents the $k$-th layer output of the search branch.
	
	\Remark Despite the weaker feature extraction capability of AlexNet compared with those deeper networks, the proposed feature transformer can make up such a drawback significantly, at the same time saving computation resources for real-time aerial tracking.
	
	\subsection{Hierarchical Feature Transformer}
	The proposed hierarchical feature transformer can be mainly divided into two steps: high-resolution features encoding and low-resolution feature decoding. The former aims at learning interdependencies between different feature layers and spatial information to raise attention to objects with different scales (especially low-resolution objects). While the latter aggregates the semantic information from the low-resolution feature map. Benefiting from the abundant global context and interdependencies among hierarchical features, our method discovers a tracking-tailored feature space. Thus, the discriminability and representative capabilities of transformed features under various aerial tracking conditions are raised significantly. Specifically, features from the last three layers are utilized. The feature map from $k$-th layer is convoluted and reshaped to $\mathbf{M}_i\in~\mathbb{R}^{WH \times C}$ ($C,~W,~H$ represents the channel, width, and height of the feature map respectively) before being fed into the feature transformer:
	\begin{equation}\label{11}
		\mathbf{M}_i=\mathcal{F}(\phi_{i}(\mathbf{Z})\star \phi_{i}(\mathbf{X}))~, i=3,4,5 ~ ,
	\end{equation}
	where $\mathcal{F}$ denotes the convolution layer and $\star$ represents the cross-correlation operator. Then, $\mathbf{M}_3'\in~\mathbb{R}^{WH\times C}$ and $\mathbf{M}_4'\in~\mathbb{R}^{WH\times C}$ can be obtained by supplementing with a learnable positional encoding.
	%In this subsection, the detail of feature-cross transformer is exposed. It can be mainly divided into two steps: the construction of feature transformation and building the multi-level structure. Considering the limited resource on UAV, we design an efficient feature transformation via low-resolution feature maps. Based on the powerful semantic information involved in the features, the feature transformation is easy to aggregate the global information for enriching the transformed features. In this way, the discriminability of transformed features is significantly raised for handling various UAV tracking conditions. Besides, to avoid the shortcoming caused by transformer, we develop a cross layer to aggregate the interdependencies between different features for better exploiting the multi-features.
	
	% For obtaining satisfying performance without sacrificing the speed, we design the feature transformer without object queries to introduce global image context information efficiently. Based on it, our tracker can maintain the stable performance under severe motion and variation conditions. In addition, to solve the shortcomings in objects with various scales, we design the multi-level structure, \textit{i.e.}, feature-cross transformer. 
	
	\subsubsection{Feature Encoding}\label{Sec:3.2.1}
	
	To fully explore the interdependencies between hierarchical features, we use the combination of $\mathbf{M}_3'$ and $\mathbf{M}_4'$ as the input of multi-head attention module~\cite{vaswani2017attention} as $\mathbf{M}_E^{1}=\mathrm{Norm}(\mathbf{M}_3'+\mathbf{M}_4')$, where $\mathrm{Norm}$ represents the normalization layer. Generally, the scaled dot-product attention $\mathrm{Att}$ can be expressed by:
	\begin{equation}
		\begin{split}
			\mathrm{Att}(\mathbf{Q},\mathbf{K},\mathbf{V})=\mathrm{Softmax}(\dfrac{\mathbf{Q}\mathbf{K}^{\mathrm{T}}}{\sqrt{c}})\mathbf{V}
		\end{split}
		~ ,
	\end{equation}
	where $\sqrt{c}$ is the scaling factor to avoid gradient vanishment in the softmax function. Then the calculation process of the multi-head attention module $\mathrm{mAtt}$ is expressed as:
	\begin{equation}\label{e1}
		\begin{split}
			&\mathrm{mAtt}(\mathbf{Q},\mathbf{K},\mathbf{V})=\Big(\mathrm{Cat}(a^1,...,a^N)\Big)\mathbf{W}_c\\
			&a^{j}=\mathrm{Att}(\mathbf{Q}\mathbf{W}^j_1,\mathbf{K}\mathbf{W}_2^j,\mathbf{V}\mathbf{W}_3^j)\\
		\end{split}
		~ ,
	\end{equation}
	where $\mathbf{W}_c\in~\mathbb{R}^{C\times C}$, $\mathbf{W}^j_1\in~\mathbb{R}^{C\times C_d}$, $\mathbf{W}^j_2\in~\mathbb{R}^{C\times C_d}$, and $\mathbf{W}^j_3\in~\mathbb{R}^{C\times C_d}$ ($C_d$=$C/N$, $N$ is the number of parallel attention head) can all be regarded as fully connected layer operation. Please note that $\mathbf{Q},\mathbf{K},\mathbf{V}$ are only mathematical symbols to clarify the function. Therefore, they do not have practical meanings. Afterwards, the output of the first multi-head attention module, \textit{i.e.}, $\mathbf{M}_E^{2}\in~\mathbb{R}^{WH\times C}$, can be obtained by:
	\begin{equation}\label{2}
		\begin{split}
			\mathbf{M}_E^{2}=\mathrm{mAtt}(\mathbf{M}_E^{1},\mathbf{M}_E^{1},\mathbf{M}_3') ~ .
		\end{split}
	\end{equation}
	
\begin{figure}[t]
		\centering
		% 调整比例，添加图片的相对位置
		%\includegraphics[scale=0.5]{images/1.pdf}
		\includegraphics[width=0.9\linewidth]{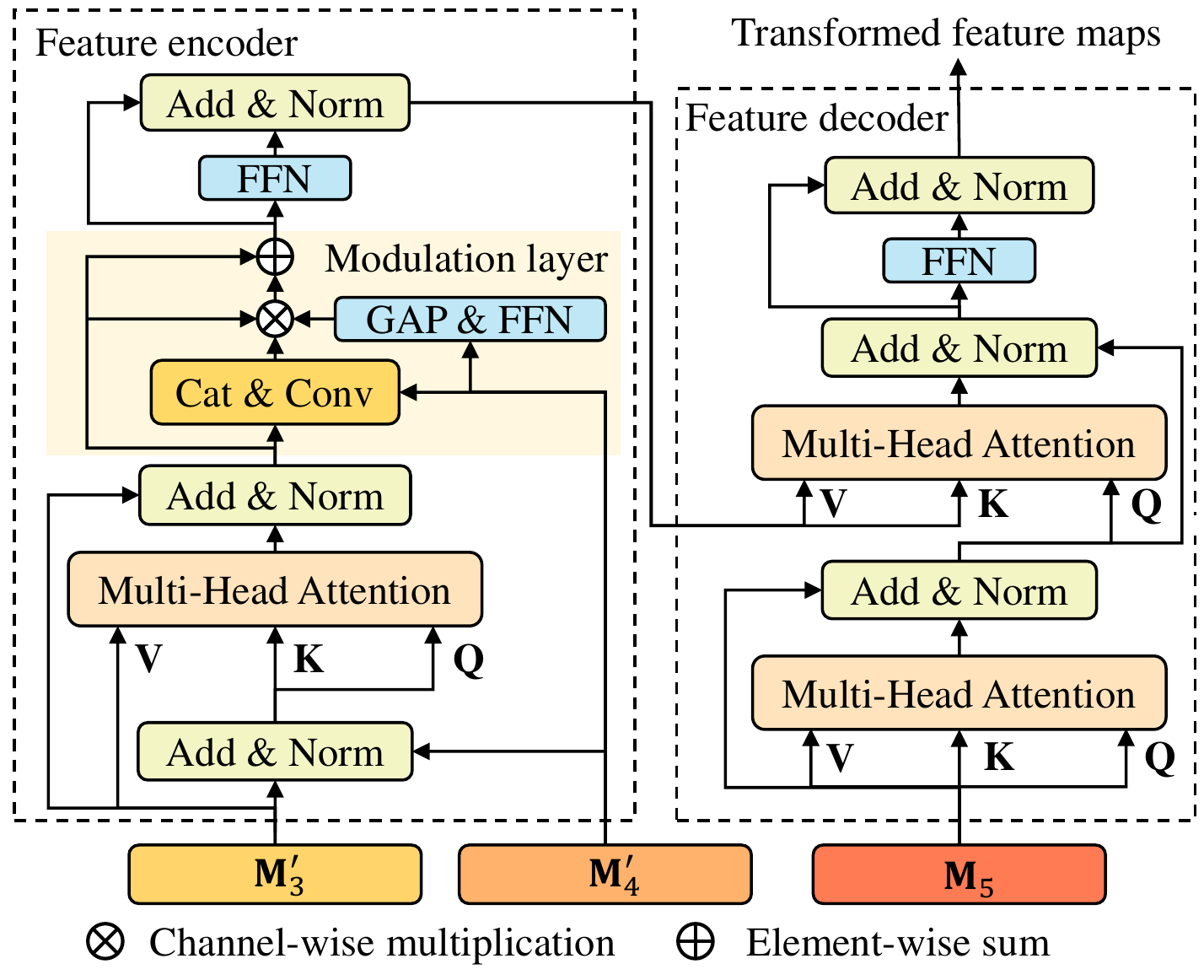}
		%\scalebox{0.5}[0.5]{\includegraphics[trim={0 75 0 55},clip]{images/1.pdf}}
		% 写标题
		\caption{Detailed workflow of HiFT. The left sub-window illustrates the feature encoder. The right one shows the structure of the decoder. Best viewed in color.
		}
		\vspace{-10pt}
		\label{fig:work} % 写 \label 跟在 \caption 后面，之后使用 \ref{}引用
	\end{figure}

	As a result, the interdependencies between $\mathbf{M}_3'$ and $\mathbf{M}_4'$ are effectively learned to enrich the high-resolution feature map $\mathbf{M}_E^{2}$. Besides, the global context in the two feature maps is also introduced in $\mathbf{M}_E^{2}$. After that, we construct the modulation layer to fully explore the potential of interdependencies between $\mathbf{M}_E^{3}$ and $\mathbf{M}_4'$ whose structure is shown in Fig.~\ref{fig:work}. Specifically, the input of modulation layer $\mathbf{M}_E^{3}$ is obtained by normalization of $\mathbf{M}_3'$ and $\mathbf{M}_E^{2}$, \textit{i.e.}, $\mathbf{M}_E^{3}=\mathrm{Norm}(\mathbf{M}_3'+\mathbf{M}_E^{2})$. After a feed-forward network (FFN) and global average pooling operation (GAP), the output of modulation layer $\mathbf{M}_E^{4}$ can be formulated as:
	\begin{equation}
		\small
		\begin{split}
			&\mathbf{W'}=\mathcal{F}(\mathrm{Cat}(\mathbf{M}_E^{3},\mathbf{M}_4'))*\mathrm{FFN}(\mathrm{GAP}(\mathbf{M}_4'))\\
			&\mathbf{M}_E^{4}=\mathbf{M}_E^{3}+\gamma_1*\mathbf{W'}*\mathbf{M}_E^{3}\\
		\end{split}
		~,
	\end{equation}
	where $\gamma_1$ represents a learning weight.
	
	Owing to the modulation layer, the internal spatial information between $\mathbf{M}_{4}'$ and $\mathbf{M}_E^{3}$ are exploited efficiently, thereby effectively distinguishing the object from the complex background. Eventually, the encoded information can be calculated through FFN and normalization.
	
	\Remark Attributing to the feature encoder, the global context and interdependencies between $\mathbf{M}_3'$ and $\mathbf{M}_4'$ are fully explored. Additionally, to overcome the deficiency of handling small objects, the modulation layer is proposed to further explore spatial information for enriching the encoded information. Finally, based on it, the decoder can build an effective feature transformation for robust tracking.

	%, where $\mathbf{M}_3'$, $\mathbf{M}_4'$, and $\mathbf{M}_5$ corresponding to the feature maps in Fig.~\ref{fig:main}
	
	\subsubsection{Feature Decoding}\label{Sec:decoder}
	Before decoding, the low-resolution feature map is firstly reshaped to $\mathbf{M}_5\in\mathbb{R}^{WH\times C}$ in Eq.~(\ref{11}). The feature decoder follows the similar structure of standard transformer~\cite{vaswani2017attention}. Differently, we build the effective feature decoder without position encoding. Since we treat the number of locations as the sequence length in our method, the position encoding is introduced to distinguish each location on feature maps. For avoiding the direct influence on the transformed feature, we decide to introduce the position information through the encoder implicitly. Analysis of the positional encoding strategy is conducted later in Sec.~\ref{Sec:abla}. The structure of the decoder is exhibited in Fig.~\ref{fig:work}.
	
	\Remark By the hierarchical feature transformer, the spatial/semantic information in the high-/low-resolution features is fully utilized to improve the discriminability of the final transformed feature. Meanwhile, the modulation layer achieves the aggregation of interdependencies among different feature layers, enhancing the robustness of tracking objects with various scales.

	\subsection{Definition of Classification Label}\label{Sec:label}

	The structures of classification and regression are implemented by several convolution layers. To achieve accurate classification, we apply two classification branches. One branch aims to classify via the area involved in the ground truth box. The other branch focuses on determining the positive samples measured by the distance between the center of ground truth and the corresponding point. Besides, to accelerate the convergence, we use pseudo-random number generators denoted as $\mathcal{T}$ to constrain the number of negative labels. 
	
	%Let denote $(g_{x},g_{y}),(g_{w},g_{h})$ as the coordinate of the center, width and height of ground truth box on the transformed feature maps.
	%supply
	%As shown in Fig.~\ref{fig:label}, the first classification label $\mathbf{P}^{cls1}\in~\mathbb{R}^{W\times H \times 1}$ can be expressed by:
	%\begin{equation}
	%\mathbf{P}^{cls1}(i,j)=
	%\left\{
	%\begin{array}{ll}
	%1, &  (i,j)~\mathrm{in}~\mathrm{R2}\\
	%-2,& (i,j)~\mathrm{in}~\complement_{\mathrm{R1}}\mathrm{R2}\\
	%0, &  (i,j)~\mathrm{in}~\mathcal{T}( \mathrm{R1} )
	%\end{array}
	%\right.
	%~ ,
	%\end{equation}
	%where $\mathrm{R1}$ and $\mathrm{R2}$ have the same center coordinates and different scales compared with ground truth. 
	
	%Besides, $\mathbf{P}^{cls2}\in~\mathbb{R}^{W\times H \times 1}$ is implemented to distinguish the closer points to ground truth based on euclidean distance between the corresponding points and the center of ground truth. Considering the core area of classification is the region around ground truth. We set the points outside of the C1 as negative samples and the inside of the C2 as positive samples. 
	
	%supply
	%Thus, the $\mathbf{P}^{cls2}$ can be written as:
	%\begin{equation}
	%\mathbf{P}^{cls2}(i,j)=
	%\left\{
	%\begin{array}{ll}
	%1, &  (i,j)~\mathrm{in}~\mathrm{C2}\\
	%Dis(i,j),& (i,j)~\mathrm{in}~\complement_{\mathrm{C1}}\mathrm{C2}\\
	%0, &  (i,j)~\mathrm{out}~\mathrm{C1}~
	%\end{array}
	%\right.
	%~ ,
	%\end{equation}
	%where $Dis(i,j)$ represents the score calculated by Euclidean distance between the point $(i,j)$ and ground truth. 
	
	\Remark The detailed calculation process of classification and regression can be found in the supplementary material.
	
	%\begin{figure}[t]
	%	\centering
	%	% 调整比例，添加图片的相对位置
	%	%\includegraphics[scale=0.5]{images/1.pdf}
	%	\includegraphics[width=0.9\linewidth]{images/label.pdf}
	%	%\scalebox{0.5}[0.5]{\includegraphics[trim={0 75 0 55},clip]{images/1.pdf}}
	%	% 写标题
	%	\caption{Visualization of classification labels $\mathbf{P}^{cls1}$ and $\mathbf{P}^{cls2}$. The C1, C2, R1, and R2 represent the circular and rectangle areas with different scales whose centers are the same as ground truth. (Best viewed in color version)}
	%	\vspace{-7pt}
	%	\label{fig:label} % 写 \label 跟在 \caption 后面，之后使用 \ref{}引用
	%\end{figure}
	
	Therefore, the overall loss function can be determined as:
	\begin{equation}
		\begin{split}
			L_{overall}=\lambda_{1}L_{cls1}+\lambda_{2}L_{cls2}+\lambda_{3}L_{loc} ~ ,
		\end{split}
	\end{equation}
	where $L_{cls1}$, $L_{cls2}$, $L_{loc}$ represent the cross-entropy, binary cross-entropy, and IoU loss. $\lambda_{1}$, $\lambda_{2}$, and $\lambda_{3}$ are the coefficients to balance the contributions of each loss.

	\section{Experiments}
	
	\subsection{Implementation Details}
	\label{subsec:EvaCri}
	During the training of 70 epochs, the last three layers of AlexNet are fine-tuned in the last 60 epochs while the first two layers are frozen. The learning rate is initialized as 5$\times 10^{-4}$ and decreased in the log space from $10^{-2}$ to $10^{-4}$. Besides, the sizes of $\mathbf{Z}$ and $\mathbf{X}$ are set to $3 \times 127 \times 127$ and   $3 \times 287 \times 287$ respectively. The feature transformer consists of one encoder layer and two decoder layers. We use image pairs extracted from COCO~\cite{lin2014microsoft}, ImageNet VID~\cite{russakovsky2015imagenet}, GOT-10K~\cite{huang2019got}, and Youtube-BB~\cite{real2017youtube} to train HiFT. In addition, the stochastic gradient descent (SGD) is adopted, and batch size, momentum, and weight decay are set to $220$, $0.9$, and $10^{-4}$, respectively. Our tracker is trained on a PC with an Intel i9-9920X CPU, a 32GB RAM, and two NVIDIA TITAN RTX GPUs.  More experimental results can be found in the supplementary.

	\subsection{Evaluation Metrics}
	The one-pass evaluation (OPE) metrics~\cite{Mueller2016ECCV} including precision and success rate are applied to assess the tracking performance. Specifically, the success rate is measured by the intersection over union (IoU) of the ground truth and estimated bounding boxes. The percentage of frames whose IoU is beyond a pre-defined threshold is drawn as the success plot (SP). Besides, the center location error (CLE) between the estimated location and the ground truth is employed to evaluate the precision. The percentage of frames whose CLE is within a certain threshold is drawn as the precision plot (PP). Meanwhile, the area under the curve (AUC) of the SP and the precision at a threshold of 20 pixels is adopted to rank the trackers.
	
	\begin{figure*}[t]
		\centering
		\subfloat[Results on DTB70.]
		{
			\includegraphics[width=0.242\linewidth]{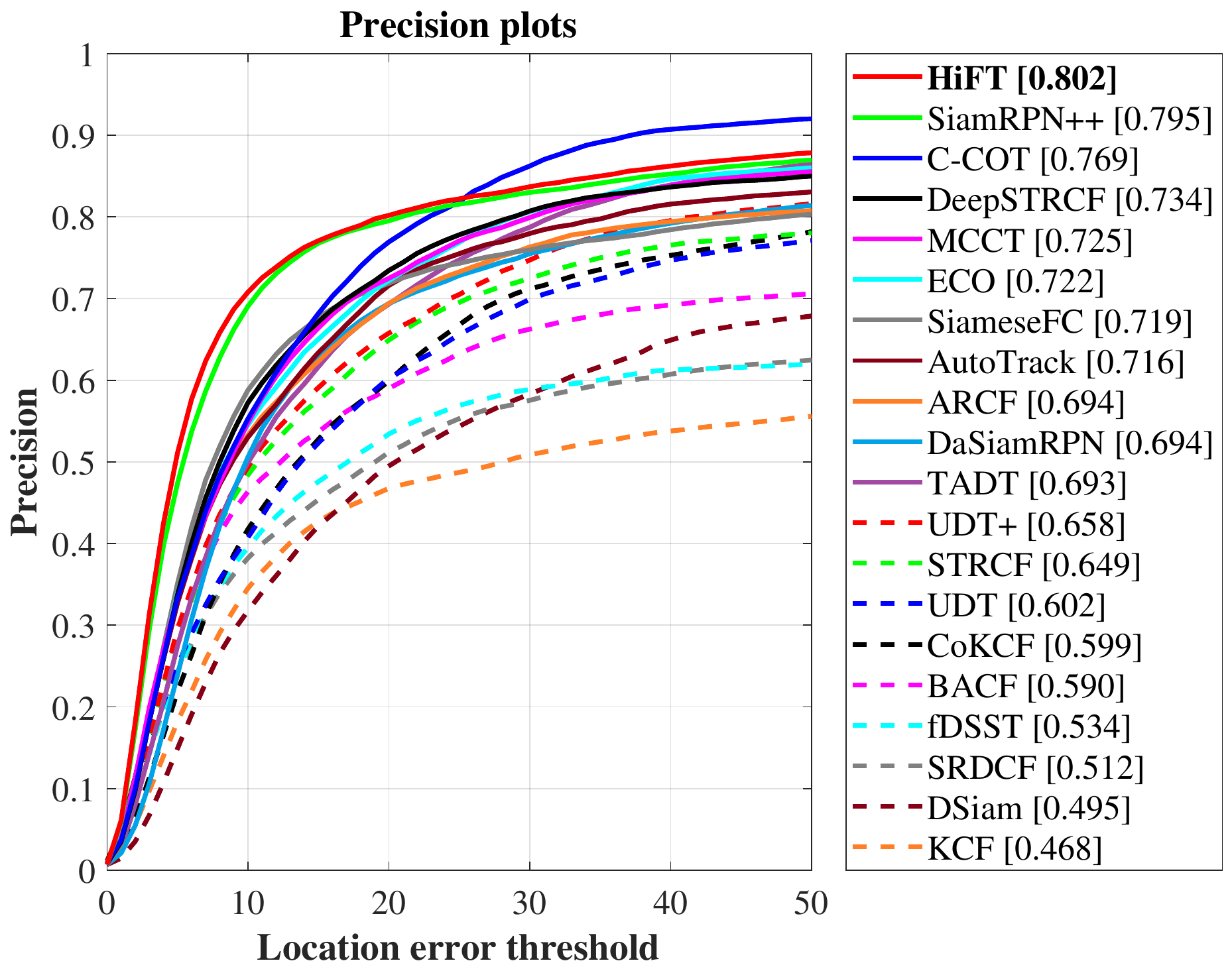}
			\includegraphics[width=0.242\linewidth]{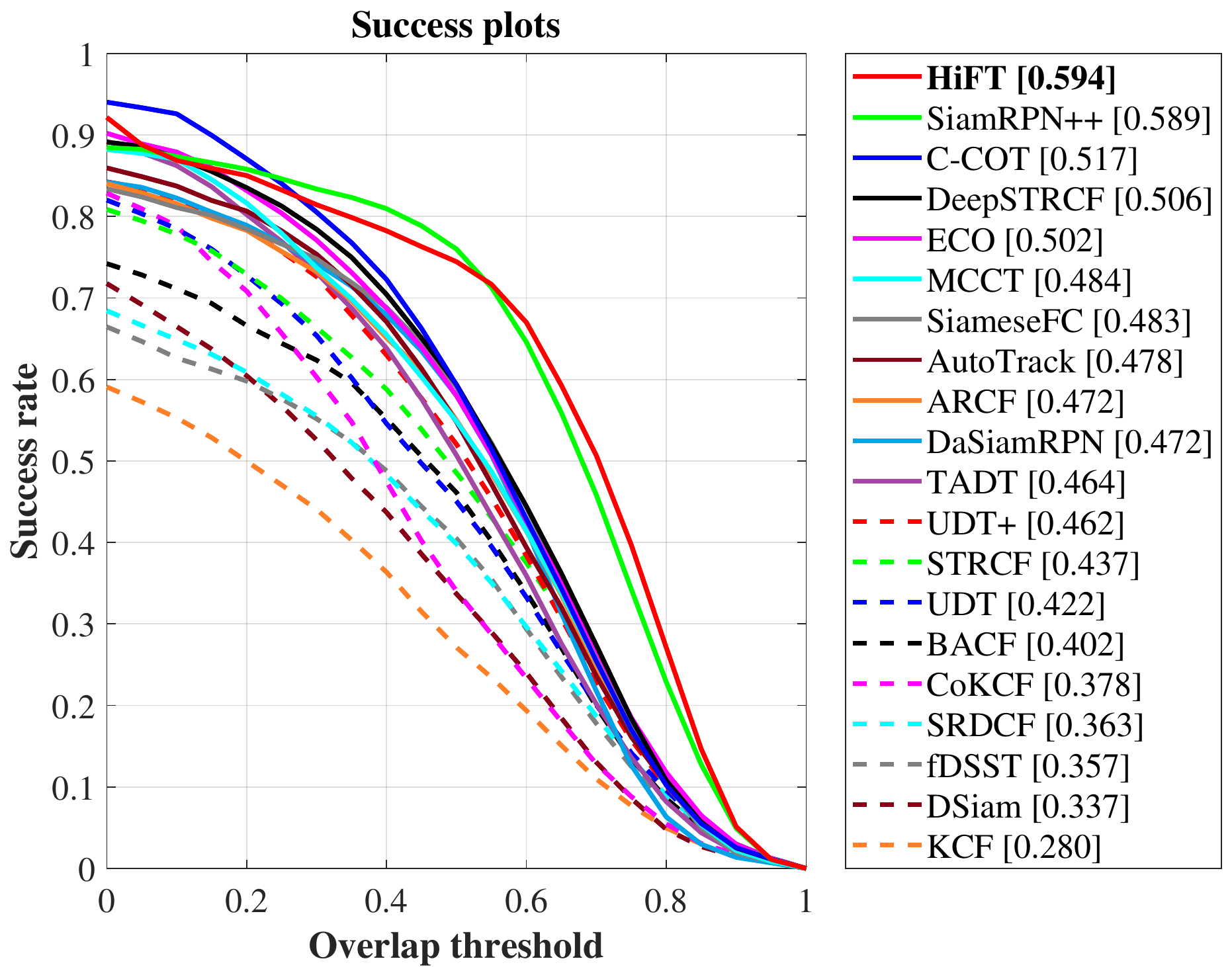}
			\label{fig:dtb70}
		}
		\subfloat[Results on UAV123@10fps.]
		{
			\includegraphics[width=0.242\linewidth]{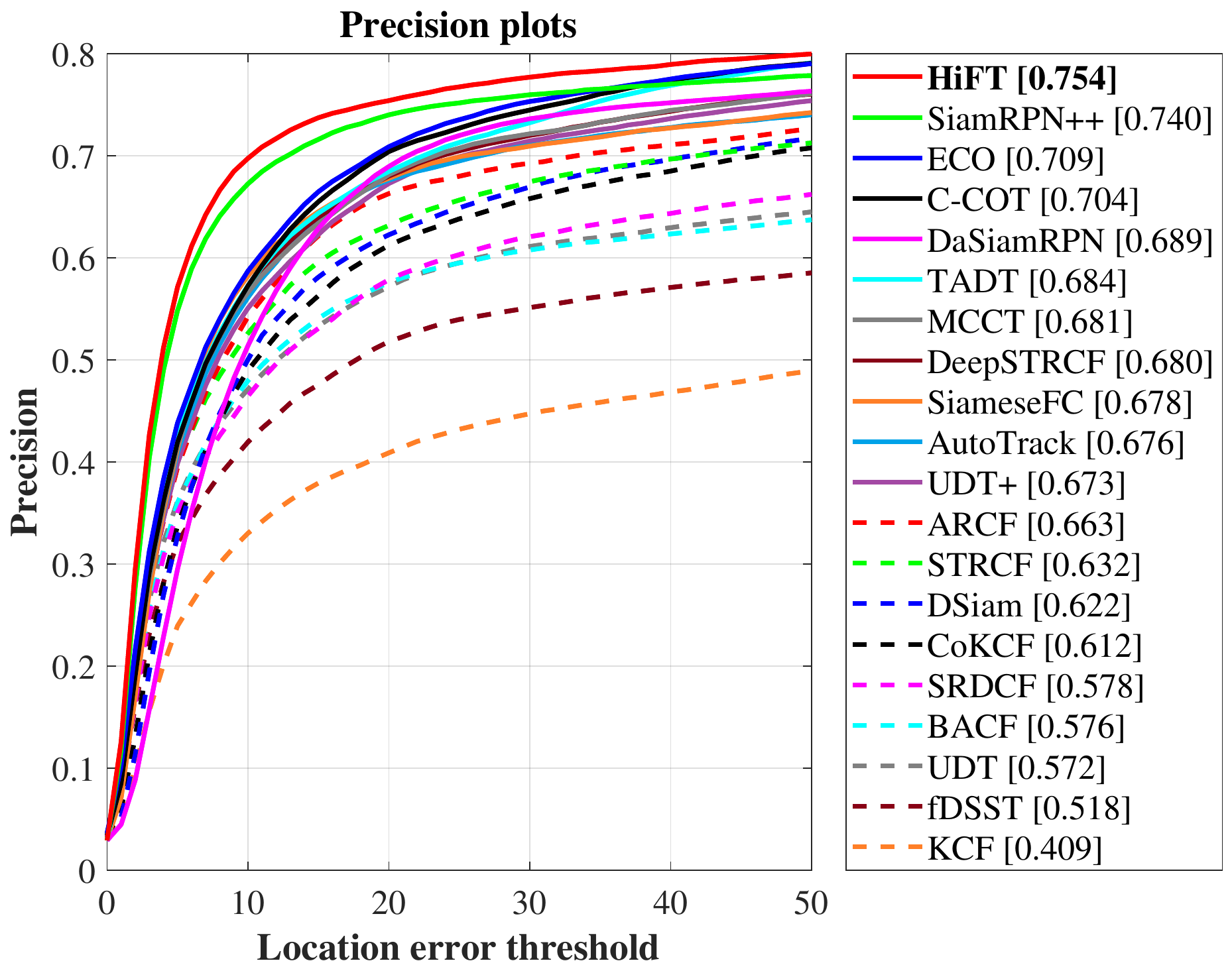}
			\includegraphics[width=0.242\linewidth]{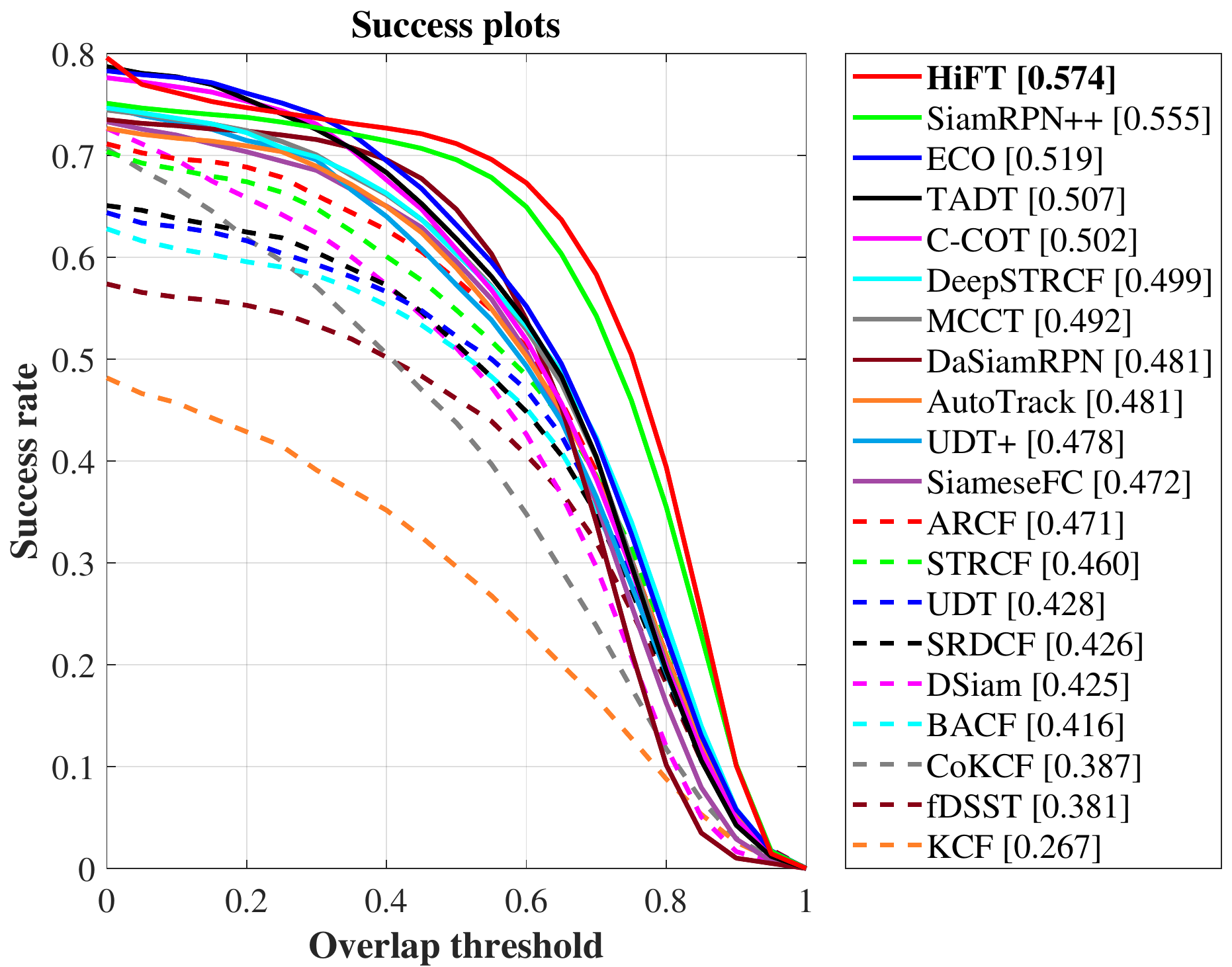}
			\label{fig:10fps}
		}
		\caption{PPs and SPs of HiFT and other SOTA trackers on (a) DTB70 and (b) UAV123@10fps. Our tracker achieves superior performance in the two benchmarks}
		
	\end{figure*}

	\subsection{Evaluation on Aerial Benchmarks}
	\subsubsection{Overall Performance}
	For overall evaluation, HiFT is tested on four challenging and authoritative aerial tracking benchmarks, and comprehensively compared with other 19 state-of-the-art (SOTA) trackers including SiamRPN++~\cite{8954116}, DaSiamRPN~\cite{zhu2018distractor}, UDT~\cite{Wang_2019_Unsupervised}, UDT+~\cite{Wang_2019_Unsupervised}, TADT~\cite{Xin2019CVPR}, CoKCF~\cite{zhang2017robust}, ARCF~\cite{huang2019learning}, AutoTrack~\cite{Li_2020_CVPR}, ECO~\cite{8100216}, C-COT~\cite{danelljan2016beyond}, MCCT~\cite{8578607}, DeepSTRCF~\cite{8578613}, STRCF~\cite{8578613}, BACF~\cite{kiani2017learning}, SRDCF~\cite{7410847}, fDSST~\cite{7569092}, SiameseFC~\cite{bertinettofully}, DSiam~\cite{guo2017learning}, and KCF~\cite{6870486}. For fairness, all the Siamese-based trackers adopt the same backbone, \textit{i.e.}, AlexNet~\cite{krizhevsky2012imagenet} pre-trained on ImageNet~\cite{russakovsky2015imagenet}.
	
	% % Table generated by Excel2LaTeX from sheet 'res_summarized'
	% \begin{table}[!b]
	%   \centering
	%   \caption{Performance on UAV20L.}
	%     \resizebox{0.95\linewidth}{!}{
	%     \begin{tabular}{lcc||lcc}
	%     \hline
	%     \hline
	%     Trackers & Prec.  & Succ.  & Trackers & Prec.  & Succ. \\
	%     \hline
	%     MCCT   & 0.605  & 0.407  & KCF    & 0.311  & 0.196 \\
	%     DSiam  & 0.603  & 0.391  & BACF   & 0.584  & 0.415 \\
	%     ECO    & 0.589  & 0.427  & DaSiamRPN & \textcolor[rgb]{ 0,  0,  1}{\textbf{0.665}} & \textcolor[rgb]{ 0,  0,  1}{\textbf{0.465}} \\
	%     STRCF  & 0.575  & 0.411  & fDSST  & 0.385  & 0.288 \\
	%     UDT    & 0.514  & 0.363  & SRDCF  & 0.507  & 0.343 \\
	%     TADT   & 0.609  & 0.459  & CoKCF  & 0.507  & 0.298 \\
	%     DeepSTRCF & 0.588  & 0.443  & AutoTrack & 0.512  & 0.349 \\
	%     C-COT  & 0.561  & 0.395  & ARCF   & 0.544  & 0.381 \\
	%     SiameseFC & 0.599  & 0.402  & SiamRPN++ & \textcolor[rgb]{ 0,  1,  0}{\textbf{0.696}} & \textcolor[rgb]{ 0,  1,  0}{\textbf{0.528}} \\
	%     UDT+   & 0.585  & 0.401  & \bf{HiFT (ours)} & \textcolor[rgb]{ 1,  0,  0}{\textbf{0.749}} & \textcolor[rgb]{ 1,  0,  0}{\textbf{0.567}} \\
	%     \hline
	%     \hline
	%     \end{tabular}}
	%   \label{tab:UAV20L}%
	% \end{table}%

	\noindent\textbf{UAV123~\cite{Mueller2016ECCV}:}
	%The proposed method, \textit{i.e.}, HiFT, obtains an impressive improvement against other SOTA trackers on four authoritative benchmarks.
	UAV123 is a large-scale UAV benchmark including 123 high-quality sequences with more than 112\textit{K} frames which covers a variety of challenging aerial scenarios such as frequent occlusion, low resolution, out-of-view, \etc. Therefore, UAV123 can help to exhaustively assess tracking performance in aerial tracking. As illustrated in Table~\ref{tab:UAV123}, HiFT outperforms other trackers in both precision and success. In terms of precision, HiFT gains first place with a precision score of 0.787, surpassing the second- and third-best SiamRPN++ (0.769) and ECO (0.752) by 2.3\% and 4.7\% respectively. As for the success rate, HiFT (0.589) also improves over SiamRPN++ (0.579) and ranks first place. In a word, HiFT demonstrates superior performance in all kinds of aerial tracking scenarios.
	\begin{table}[!b]
		\caption{Quantitative evluation on UAV123~\cite{Mueller2016ECCV}. The top three
			performances are respectively highlighted by \textcolor[rgb]{ 1,  0,  0}{\textbf{red}}, \textcolor[rgb]{ 0,  1,  0}{\textbf{green}}, and \textcolor[rgb]{ 0,  0,  1}{\textbf{blue}} color. Prec. and Succ. respectively denote precision score at 20 pixels and AUC of success plot.}
		\vspace{4pt}
		\centering
		\renewcommand\tabcolsep{8pt}
		\resizebox{0.99\linewidth}{!}{
			\begin{tabular}{lcc|lcc}
				\toprule
				Trackers & Prec.  & Succ.  & Trackers & Prec.  & Succ. \\
				\midrule
				AutoTrack~\cite{Li_2020_CVPR} & 0.689  & 0.472  & C-COT~\cite{danelljan2016beyond} & 0.729  & 0.502 \\
				ARCF~\cite{huang2019learning} & 0.671  & 0.468  & 
				UDT+~\cite{Wang_2019_Unsupervised} & 0.732  & 0.502  \\
				STRCF~\cite{8578613} & 0.681  & 0.481  & UDT~\cite{Wang_2019_Unsupervised} & 0.668  & 0.477 \\
				fDSST~\cite{7569092} & 0.583  & 0.405  & TADT~\cite{Xin2019CVPR} & 0.727  & 0.520 \\
				SRDCF~\cite{7410847} & 0.676  & 0.463  & DeepSTRCF~\cite{8578613} & 0.705  & 0.508 \\
				CoKCF~\cite{zhang2017robust} & 0.652  & 0.399  & MCCT~\cite{8578607} & 0.734  & 0.507 \\
				KCF~\cite{6870486} & 0.523  & 0.331  & DSiam~\cite{guo2017learning} & 0.608  & 0.400 \\
				BACF~\cite{kiani2017learning} & 0.662  & 0.461  & ECO~\cite{8100216} & \textcolor[rgb]{ 0,  0,  1}{\textbf{0.752}} & \textcolor[rgb]{ 0,  0,  1}{\textbf{0.528}} \\
				SiamRPN++~\cite{8954116} & \textcolor[rgb]{ 0,  1,  0}{\textbf{0.769}} & \textcolor[rgb]{ 0,  1,  0}{\textbf{0.579}} & SiameseFC~\cite{bertinettofully} & 0.725  & 0.494\\
				DaSiamRPN~\cite{zhu2018distractor} & 0.725  & 0.501  & \textbf{HiFT (ours)} & \textcolor[rgb]{ 1,  0,  0}{\textbf{0.787}} & \textcolor[rgb]{ 1,  0,  0}{\textbf{0.589}} \\
				\bottomrule
		\end{tabular}}
		\label{tab:UAV123}
	\end{table}%
	% Table generated by Excel2LaTeX from sheet 'res_summarized'
	\begin{table}[!b]
		\centering
		\caption{Overall evaluation on UAV20L~\cite{Mueller2016ECCV}. The top nine trackers are reported.The top three
			trackers are respectively marked by \textcolor[rgb]{ 1,  0,  0}{\textbf{red}}, \textcolor[rgb]{ 0,  1,  0}{\textbf{green}}, and \textcolor[rgb]{ 0,  0,  1}{\textbf{blue}} color. Prec. and Succ. respectively denote precision score at 20 pixels and AUC of success plot. }
		\vspace{4pt}
		\renewcommand\tabcolsep{2pt}
		\resizebox{0.99\linewidth}{!}{
			\begin{tabular}{cccccccccc}
				\toprule
				& UDT+  & ECO    & TADT   & DeepST- & Siames- & DSiam   & DaSiam & SiamRP- & \bf{HiFT} \\
				& \cite{Wang_2019_Unsupervised} & \cite{8100216} & \cite{Xin2019CVPR} & RCF~\cite{8578613} & eFC~\cite{bertinettofully} & \cite{guo2017learning} & RPN~\cite{zhu2018distractor} & N++~\cite{8954116} &  \bf{(ours)} \\
				\midrule
				Prec.  & 0.585  & 0.589  & 0.609  & 0.588  & 0.599  & 0.603  & \textcolor[rgb]{ 0,  0,  1}{\textbf{0.665}} & \textcolor[rgb]{ 0,  1,  0}{\textbf{0.696}} & \textcolor[rgb]{ 1,  0,  0}{\textbf{0.763}} \\
				Succ.  & 0.401  & 0.427  & 0.459  & 0.443  & 0.402  & 0.391  & \textcolor[rgb]{ 0,  0,  1}{\textbf{0.465}} & \textcolor[rgb]{ 0,  1,  0}{\textbf{0.528}} & \textcolor[rgb]{ 1,  0,  0}{\textbf{0.566}} \\
				\bottomrule
		\end{tabular}}
		\label{tab:uav20l}%
	\end{table}%
	\begin{table*}[t]
		\footnotesize
		\setlength{\tabcolsep}{3mm}
		\centering
		\caption{Average evaluation on four aerial tracking benchmarks.  Our tracker outperforms all other trackers with an obvious improvement. The best three
			performances are respectively highlighted with \textcolor[rgb]{ 1,  0,  0}{\textbf{red}}, \textcolor[rgb]{ 0,  1,  0}{\textbf{green}}, and \textcolor[rgb]{ 0,  0,  1}{\textbf{blue}} color.}
		\vspace{4pt}
		\centering
		\renewcommand\tabcolsep{3pt}
		\resizebox{0.95\linewidth}{!}
		{
			\begin{tabular}{ccccccccccccc}
				\toprule
				\multirow{2}[1]{*}{Trackers} & \bf{HiFT}& SiamRPN++ & DaSiamRPN&AutoTrack & ARCF    & C-COT  & SiameseFC & UDT+   & TADT   & DeepSTRCF & MCCT   & ECO     \\
				& (ours)& \cite{8954116} & \cite{zhu2018distractor}&\cite{Li_2020_CVPR} & \cite{huang2019learning}  & \cite{danelljan2016beyond} & \cite{bertinettofully} & \cite{Wang_2019_Unsupervised} & \cite{Xin2019CVPR} & \cite{8578613} & \cite{8578607} & \cite{8100216}  \\
				\midrule
				Avg. Prec. &\textcolor[rgb]{ 1,  0,  0}{\textbf{0.776}}& \textcolor[rgb]{ 0,  1,  0}{\textbf{0.750}} & \textcolor[rgb]{ 0,  0,  1}{\textbf{0.693}}& 0.648   & 0.643   & 0.691   & 0.680   & 0.662   & 0.678   & 0.677   & 0.686   & 0.693    \\
				Avg. Succ. & \textcolor[rgb]{ 1,  0,  0}{\textbf{0.581}}& \textcolor[rgb]{ 0,  1,  0}{\textbf{0.563}} & 0.480 & 0.445   & 0.448    & 0.479   & 0.463   & 0.461   & 0.488   & 0.489   & 0.472   & \textcolor[rgb]{ 0,  0,  1}{\textbf{0.494}}  \\
				\bottomrule
			\end{tabular}%
		}
		\label{tab:over}%
	\end{table*}%

	\noindent\textbf{UAV20L~\cite{Mueller2016ECCV}:}
	UAV20L is composed of 20 long-term tracking sequences with 2934 frames on average and over 58$K$ frames in total. In this paper, it is utilized to evaluate our tracker in realistic long-term aerial tracking scenes. As presented in Table~\ref{tab:uav20l}, attributing to the global contextual information introduced by the feature transformer, our tracker achieves competitive performance compared to other SOTA trackers. Specifically, HiFT yields the best precision score (0.763), surpassing the second-best SiamRPN++ (0.696) and the third-best DaSiamRPN (0.665) by \textbf{9.6\%} and \textbf{14.7\%}. Similarly, in success rate, HiFT achieves the best score (0.566), followed by SiamRPN++ (0.528) and DaSiamRPN (0.465). The extraordinary performance verifies that HiFT 
	could be a desirable choice in long-term aerial tracking scenarios.
	
	% Table generated by Excel2LaTeX from sheet 'res_summarized'

	% \begin{figure}[t]
	% \centering
	% \includegraphics[width=0.49\linewidth]{res/UAV123@10fps/figs_OPE/error_OPE.pdf}
	% \includegraphics[width=0.49\linewidth]{res/UAV123@10fps/figs_OPE/overlap_OPE.pdf}
	% \caption{Overall performance of UAV123@10fps.}
	% \end{figure}

	\noindent\textbf{DTB70~\cite{li2017visual}:} 
	Compared to the aforementioned two benchmarks, DTB70 contains 70 challenging UAV sequences with a large number of severe motion scenes. The robustness of trackers in fast motion scenarios could be appropriately evaluated on this benchmark. Experimental results are shown in Fig.~\ref{fig:dtb70}, HiFT ranks first place in both precision (0.802) and success rate (0.594), followed by SiamRPN++ with a precision of 0.795 and a success rate of 0.589. The promising ability of HiFT in handling fast motion can be attributed to the proposed hierarchical feature transformer which is able to promote the discrimination ability of HiFT.
	
	%\begin{table*}[t]
	%	\footnotesize
	%	\setlength{\tabcolsep}{2mm}
	%	\centering
	%	\caption{Comparison other SOTA tracker equipped with different backbone on DTB70 benchmarks. Note that RN and MN represent the Resnet and mobilenetv2 backbone. The third line reflects the differences between other trackers and the HiFT. The best three performances are respectively highlighted with \NoOne{\textbf{red}}, \NoTwo{\textbf{green}}, and \NoThree{\textbf{blue}} color.}
	%	\vspace{4pt}
	%	\centering
	%	\setlength{\tabcolsep}{0.8mm}
	%	{
	%		\begin{tabular}{l ccccccccccc}
	%			\hline
	%			\hline
	%			{\textbf{Trackers}}&{\textbf{HiFT}}&{SiamRPN++\_RN}&{SiamMask\_RN}&{SiamRPN++\_MN}&ATOM&PrDiMP\_RN50&SiamCAR\_RN&DiMP\_RN50&SiamBAN\_RN\\  
	%			\hline
	%			\textbf{Precision (\%)} & \NoThree{\textbf{80.2}}&80.0&77.1&78.7&76.7&75.9&77.4&\NoOne{\textbf{82.0}}&\NoTwo{\textbf{81.2}}
	%			\\ 
	%			\textbf{$\Delta_{Prec}$ (\%)}& -- & 0.25 &\NoThree{\textbf{3.9}}& 1.9& \NoTwo{\textbf{4.4}}& \NoOne{\textbf{5.4}}& 3.5& -2.2  &-1.2\\ 
	%			\textbf{FPS} & \NoOne{\textbf{140}}&78&76&89&64&34&71&32&73 \\
	%			\textbf{$\Delta_{FPS}$ (\%)} & -- & 200& 200& 200& 200& 200& 200& 200& 200\\
	%			\hline
	%			\hline
	%		\end{tabular}
	%	}
	%	\label{tab:fps}%
	%\end{table*}%
	% 	\label{fig:overperform}
	% \end{figure*}

	\noindent\textbf{UAV123@10fps~\cite{Mueller2016ECCV}:} 
	UAV123@10fps is created by down-sampling from the original 30FPS recording. Consequently, the issue of strong motion in UAV123@10fps is more severe compared to UAV123. The PPs and SPs shown in Fig.~\ref{fig:10fps} demonstrate that HiFT can consistently obtain satisfactory performance, achieving the best precision (0.754) and success rate (0.574). To sum up, HiFT provides a more stable performance comparing to other SOTA trackers, verifying its favorable robustness in various aerial tracking scenarios.
	\begin{table}[b]
		
		\centering
		\caption{Attribute-based evaluation of top 6 trackers on four benchmarks. The best two
			performances are respectively highlighted by \textcolor[rgb]{ 1,  0,  0}{\textbf{red}} and \textcolor[rgb]{ 0,  1,  0}{\textbf{green}} color. HiFT keeps achieving the best performance in different attributes. $\Delta$ denotes the improvement in comparison with the second best tracker.}
		\vspace{4pt}
		\renewcommand\tabcolsep{5pt}
		\resizebox{0.99\linewidth}{!}{
			\begin{tabular}{lcccccccc}
				\toprule
				Attributes & \multicolumn{2}{c}{Low-resolution} & \multicolumn{2}{c}{Scale variation} & \multicolumn{2}{c}{Occlusion} & \multicolumn{2}{c}{Fast motion} \\
				\midrule
				Trackers & Prec.  & Succ.  & Prec.  & Succ.  & Prec.  & Succ.  & Prec.  & Succ. \\
				\midrule
				SiamRPN++ & 0.591  & \textcolor[rgb]{ 0,  1,  0}{\textbf{0.390}} & \textcolor[rgb]{ 0,  1,  0}{\textbf{0.728}}  & \textcolor[rgb]{ 0,  1,  0}{\textbf{0.559}}  & \textcolor[rgb]{ 0,  1,  0}{\textbf{0.601}}  & \textcolor[rgb]{ 0,  1,  0}{\textbf{0.405}}  & \textcolor[rgb]{ 0,  1,  0}{\textbf{0.680}}  & \textcolor[rgb]{ 0,  1,  0}{\textbf{0.489}} \\
				DaSiamRPN & 0.592  & 0.347  & 0.678  & 0.482  & 0.583  & 0.361  & 0.617  & 0.409 \\
				C-COT  & 0.586  & 0.331  & 0.643  & 0.451  & 0.571  & 0.359  & 0.644  & 0.411 \\
				TADT   & \textcolor[rgb]{ 0,  1,  0}{\textbf{0.604}}  & 0.366  & 0.632  & 0.466  & 0.598  & 0.387  & 0.628  & 0.412 \\
				ECO    & 0.581  & 0.343  & 0.644  & 0.471  & 0.583  & 0.375  & 0.620  & 0.407 \\
				\bf{HiFT (ours)} &\textcolor[rgb]{ 1,  0,  0}{\textbf{0.626}} & \textcolor[rgb]{ 1,  0,  0}{\textbf{0.416}} & \textcolor[rgb]{ 1,  0,  0}{\textbf{0.772}} & \textcolor[rgb]{ 1,  0,  0}{\textbf{0.584}} & \textcolor[rgb]{ 1,  0,  0}{\textbf{0.638}} & \textcolor[rgb]{ 1,  0,  0}{\textbf{0.431}} & \textcolor[rgb]{ 1,  0,  0}{\textbf{0.751}} & \textcolor[rgb]{ 1,  0,  0}{\textbf{0.537}} \\
				\midrule
				$\Delta$ (\%) & 3.63   & 6.81   & 5.98   & 4.40   & 6.20   & 6.43   & 10.42  & 9.79 \\
				\bottomrule
		\end{tabular}}
		\label{tab:att}%
		% 	\vspace{-10pt}
	\end{table}%

	\Remark Table~\ref{tab:over} reports the average precision and success rate of the top 11 trackers on four benchmarks. It shows that HiFT has improved the second-best tracker SiamRPN++ by 3.5\% and 3.2\% in precision and success rate respectively. 
	
	%The overall performance of HiFT surpasses the second-best tracker by over 3.2\%, which indicates that HiFT can provide robust performance for aerial tracking. %Besides, some qualitative comparisons among top 6 trackers in overall performance are shown in Fig.~\ref{fig:res}. The main challenges of these sequences include scale variation, aspect ratio change, fast motion, occlusion, camera motion, \textit{etc}. HiFT maintains a more precise estimation of object state during the challenging scenes comparing to other trackers.

	% \begin{figure}[t]
	% 	\centering	
	% 	\includegraphics[width=0.99\linewidth]{images/visualization.pdf}
	
	% 	\caption{Screenshots of \textit{BMX4, RaceCar} from DTB70, \textit{group3\_2, and person8\_1} from UAV123. More visualizations and video are released in supplementary material.} 
	
	% 	\label{fig:res}
	% \end{figure}

	\subsubsection{Attribute-based Comparison}
	\label{subsec:attr}
	To exhaustively evaluate HiFT under various challenges, attribute-based comparisons are conducted, seen in Table~\ref{tab:att}. HiFT ranks first place in terms of both precision and success rate in comparison with other top 5 trackers. Specifically, HiFT significantly exceeds the second-best performance in attributes of low-resolution, scale variation, occlusion, and fast motion. HiFT improves the second-best performance by around \textbf{10}\% in fast motion scenarios. The satisfactory results demonstrate that our hierarchical feature transformer can help exploit the global contextual information to overcome issues of severe motion. In addition, when the objects are severely occluded, HiFT can learn more robust features to discriminate the occluded objects. Therefore, HiFT also yields prominent improvement in the scenarios of occlusion. Moreover, since the multi-scale feature maps are utilized for building the feature transformation, our tracker is endowed with the ability to track objects with various scales, as verified by its performance in the attributes of low-resolution and scale variation.

	\subsubsection{Ablation Study}\label{Sec:abla}
	To verify the effectiveness of each module of the proposed method, detailed studies amongst HiFT with different modules enabled are conducted on UAV20L.
	
	\noindent\textbf{Symbol introduction:}
	For clarity, we first introduce the meaning of symbols used in Table~\ref{tab:a}. This work considers \texttt{Baseline} as the model with only feature extraction and regression \& classification network. \texttt{OT} denotes original standard transformer (with object query). \texttt{FT} indicates the original transformer with the feature map (instead of object query) but without the proposed modulation layer. \texttt{HFT} denotes the full version of the proposed hierarchical feature transformer. \texttt{PE} represents direct positional encoding to $\mathbf{M}_5$ (HiFT leaves out position encoding in $\mathbf{M}_5$ as demonstrated in Sec.~\ref{Sec:decoder}). \texttt{RL} represents the rectangle label used in the traditional trackers. For fairness, each version of the tracker adopts the same training strategy except for the investigated module. 
	\begin{table}[t]
		
		\centering
		\caption{Ablation study of different components of HiFT. For the detailed explanation of Baseline, OT, FT, HFT, PE, and RL, please kindly refer to the text in Sec.~\ref{Sec:abla}. $\Delta$ denotes the improvement compared with the Baseline tracker.}
		% 	\vspace{4pt}
		\renewcommand\tabcolsep{4pt}
		\resizebox{1.0\linewidth}{!}{
			\begin{tabular}{lcccc}
				\toprule
				
				Trackers & Precision & $\Delta_{pre}$ (\%) &Success&  $\Delta_{suc}$ (\%) \\
				\midrule
				Baseline   & 0.611 & -- & 0.463 & -- \\
				Baseline+OT   & 0.597 & -2.29 & 0.446& -3.67 \\
				
				Baseline+FT   & 0.675 & {+10.47} & 0.496 & {+7.13} \\
				\midrule
				Baseline+HFT+PE   & 0.689 & {+12.77} & 0.523 & {+12.96}\\
				Baseline+HFT+RL & 0.629 & {+2.95} & 0.486 & {+4.97} \\

				\midrule
				\textbf{Baseline+HFT (HiFT)} & \textbf{0.763} & \textbf{+24.88} & \textbf{0.566} & \textbf{+22.25} \\
				\bottomrule
		\end{tabular}}
		\label{tab:a}%
		\vspace{-10pt}
	\end{table}%
	
	\begin{figure}[b]
		\centering
		% 调整比例，添加图片的相对位置
		%\includegraphics[scale=0.5]{images/1.pdf}
		\includegraphics[width=1\linewidth]{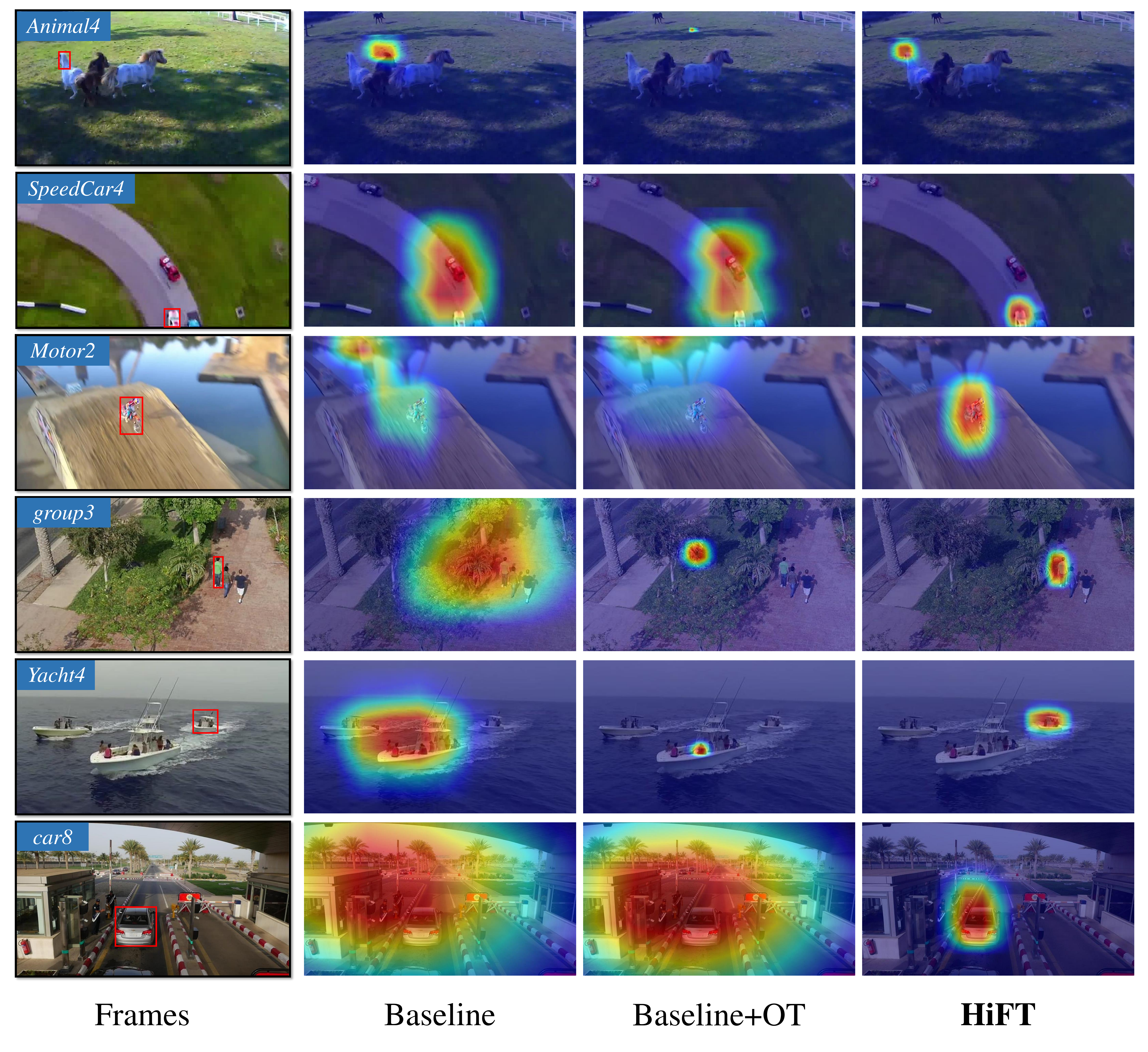}
		%\scalebox{0.5}[0.5]{\includegraphics[trim={0 75 0 55},clip]{images/1.pdf}}
		% 写标题
		\caption{Visualization of the confidence map of three tracking methods on several sequences from UAV20L~\cite{Mueller2016ECCV} and DTB70~\cite{li2017visual}. The target objects are marked out by \textbf{\textcolor{red}{red}} boxes in the original frames. HiFT gets more robust performance for visual tracking in the air.}
		\label{fig:visdeep} % 写 \label 跟在 \caption 后面，之后使用 \ref{}引用
		% 	\vspace{-14pt}
	\end{figure}

	\noindent\textbf{Discussion on transformer architecture:}
	As shown in Table~\ref{tab:a}, adding the original transformer with object queries (\texttt{Baseline+OT}) directly lowers the performance of \texttt{Baseline} by about 2.29\% on precision and 3.67\% on success rate, which proves that object queries hardly perform well in SOT with novel target objects. Replacing object query with the feature map, \texttt{Baseline+FT} raises tracking precision by \textbf{10.47\%}. Further adopting the modulation layer, \texttt{Baseline+HFT}, yields the best performance by \textbf{24.88\%}. All the aforementioned results can be combined together to validate the efficacy of the elaborately designed hierarchical feature transformer with the modulation layer in aerial tracking.
	
	\noindent\textbf{Discussion on position encoding\&classification label:}
	This part aims at proving the 2 strategies, position encoding in Sec.~\ref{Sec:decoder} and new classification label in Sec.~\ref{Sec:label}. For position decoding, in Table~\ref{tab:a}, the tracker \texttt{Baseline+HFT+PE} hurts the performance of \texttt{HiFT} tremendously (from \textbf{24.88\%} improvements to \textbf{12.77\%}), proving that direct position encoding is indeed not proper for feature $\mathbf{M}_5$. Considering the distance of ground truth and sample points, the circular strategy utilized in \texttt{HiFT} achieves a notable improvement (\textbf{24.88\%}) compared to the traditional rectangle label in \texttt{Baseline+HFT+RL} (\textbf{2.95\%}).
	
	\Remark Please note that more ablation studies are reported in supplement material.
	%Since classification plays a significant role in Siamses trackers, the classification label needs to be designed carefully. By considering the distance of ground truth and sample points, we utilize the circular strategy. The comparison against the rectangle label proves the validity of the circular classification strategy, directly demonstrating the assistance of distance information to tracking performance. Moreover, the studies about position encoding are performed for demonstrating the influence of directly introducing position encoding. The low-resolution features are calculated by backbone which is pre-trained on numerous images and refined on train dataset while the position encoding is only trained on train dataset. Therefore, due to the difference in training data, directly introducing the position information into low-resolution features is inappropriate. Table~\ref{tab:a} clearly shows the negative influence of directly introduction of position encoding, \textit{i.e.}, decreasing about \textbf{10.74}\% on precision and \textbf{8.22\%} on success rate.
	\subsubsection{Qualitative Evaluation}
	As shown in Fig.~{\ref{fig:visdeep}}, the confidence map of our HiFT tracker consistently focuses on the object under onerous challenges in aerial tracking, \textit{e.g.}, fast motion in \textit{Motor2}, low resolution in \textit{SpeedCar4}, and occlusion in \textit{group3} and \textit{Yacht4}. Despite that the \texttt{Baseline} and \texttt{Baseline+OT} are trained with the same strategy as HiFT, they still fail to concentrate on the target object in those complex tracking scenarios, which proves the robustness of the proposed hierarchical feature transformer.
	
	%Proposed HiFT maintains high robustness under various challenges like similar object (\textit{Animal4}), low resolution (\textit{SpeedCar4}), fast motion (\textit{Motor2}), occlusion (\textit{Yacht4}, \textit{group3}), and scale variation (\textit{car8}). While the tracker without transformer (\texttt{Baseline}) and with original transformer (\texttt{Baseline+OT}) fall invalid, despite the same training procedure.

	\begin{figure*}
		\centering
		\includegraphics[width=\linewidth]{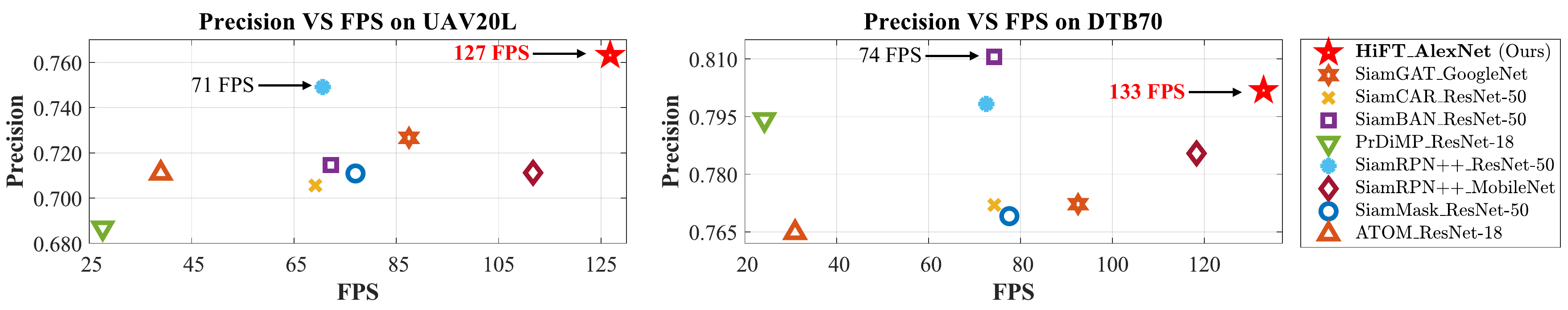}
		\caption{Precision-speed trade-off analysis by quantitative comparison between HiFT and trackers with deeper backbone on UAV20L~\cite{Mueller2016ECCV} (left) and DTB70~\cite{li2017visual} (right). Our method realizes an excellent trade-off on both two benchmarks.}
		\label{fig:deeper_star}
	\end{figure*}
	
	% Table generated by Excel2LaTeX from sheet 'res_summarized'
	\begin{table*}[!t]
		\centering
		\caption{Average precision and tracking speed of HiFT and the trackers with deeper backbone. The proposed approach runs at a satisfactory speed of $\sim$130 FPS, while achieving comparable tracking performance with those trackers equipped with a deeper backbone. The best three
			performances are respectively highlighted with \textcolor[rgb]{ 1,  0,  0}{\textbf{red}}, \textcolor[rgb]{ 0,  1,  0}{\textbf{green}}, and \textcolor[rgb]{ 0,  0,  1}{\textbf{blue}} color.}
		\resizebox{0.99\linewidth}{!}
		{
			\begin{tabular}{lccccccccc}
				\toprule
				Tracker & \bf{HiFT (ours)} & SiamGAT~\cite{Guo_2021_CVPR} & SiamCAR~\cite{9157720} & SiamBAN~\cite{chen2020siamese} & PrDiMP~\cite{9157124} & SiamRPN++~\cite{8954116} & SiamRPN++~\cite{8954116} & SiamMask~\cite{8953931} & ATOM~\cite{8953466} \\
				\midrule
				Backbone & AlexNet & GoogleNet & ResNet-50 & ResNet-50 & ResNet-18 & ResNet-50 & MobileNet & ResNet-50 & ResNet-18 \\
				Avg. Prec. & \textcolor[rgb]{ 1,  0,  0}{\textbf{0.783}} & 0.751  & 0.739  & \textcolor[rgb]{ 0,  0,  1}{\textbf{0.763}}  & 0.741  & \textcolor[rgb]{ 0,  1,  0}{\textbf{0.774}}  & 0.748  & 0.740  & 0.738 \\
				Avg. FPS    & \textcolor[rgb]{ 1,  0,  0}{\textbf{129.87}} & \textcolor[rgb]{ 0,  0,  1}{\textbf{90.01}}  & 71.74  & 73.25  & 25.94  & 71.59  & \textcolor[rgb]{ 0,  1,  0}{\textbf{115.03}} & 77.30  & 34.94 \\
				\bottomrule
		\end{tabular}}
		\vspace{-8pt}
		\label{tab:deeper_avg}%
	\end{table*}%
	
	%Owing to the extracted global context information, the tracker is skilled in handling the local drastic motion. Meanwhile, the multi-scale feature maps enhance the performance in determining the object with various scales. Most notably, HiFT maintains robustness in LR conditions with an improvement of \textbf{10\%}. 
	
	% precision of SiamAPN++ with different components and the baseline SiamAPN on UAV20L is listed in TABLE~\ref{tab:parameter}. With the internal relationship introduced by the APN-DF, the tracker has surpassed the baseline. Besides, it indeed promotes performance when tracking objects with various scales. Furthermore, attributing to the AAN, the self-interdependencies from the single feature map and the cross-interdependencies are aggregated, further improving the accuracy of SiamAPN++.

	%\subsubsection{Qualitative Evaluation}
	%Some qualitative comparisons are shown in Fig. \ref{fig:res}. The main challenges of these sequences include SV, ARC, FM, POC, CM, and OV. It clearly shows the superior performance of HiFT.

	\subsubsection{Comparison to Trackers with Deeper Backbone}
	The proposed hierarchical feature transformer dedicates to model effective feature mapping among multi-level features, so as to achieve SOTA performance without introducing a huge computational burden.  To further evaluate its effectiveness, we employ the trackers equipped with deeper backbones for comparison. The state-of-the-art trackers, including SiamRPN++ (ResNet-50)~\cite{8954116}, SiamRPN++ (MobileNet)~\cite{8954116}, SiamMask (ResNet-50)~\cite{8953931}, ATOM (ResNet-18)~\cite{8953466}, DiMP (ResNet-50)~\cite{9010649}, PrDiMP (ResNet-18)~\cite{9157124}, SiamCAR (ResNet-50)~\cite{9157720}, SiamGAT (GoogleNet)~\cite{Guo_2021_CVPR}, and SiamBAN (ResNet-50)~\cite{chen2020siamese}, are involved in the comparison. As illustrated in Fig.~\ref{fig:deeper_star}, HiFT achieves a satisfactory balance of tracking robustness and speed. On UAV20L, adopting AlexNet as the backbone, HiFT (0.763) surpasses the second-best tracker SiamRPN++\_ResNet-50 (0.749) in precision and achieves a speed of 127 FPS, which is \textbf{1.8} times faster than the latter. Similarly, on DTB70, HiFT achieves comparable performance compared to those deeper CNN-based trackers. Eventually, the average precision and tracking speed are reported in Table~\ref{tab:deeper_avg}, HiFT yields the best average precision (0.783) with a promising speed of \textbf{129.87} FPS, proving that HiFT achieves an excellent balance between tracking performance and efficiency.

	\section{Real-World Tests}
	
\begin{figure}[!t]
		\centering	
		\includegraphics[width=1\linewidth]{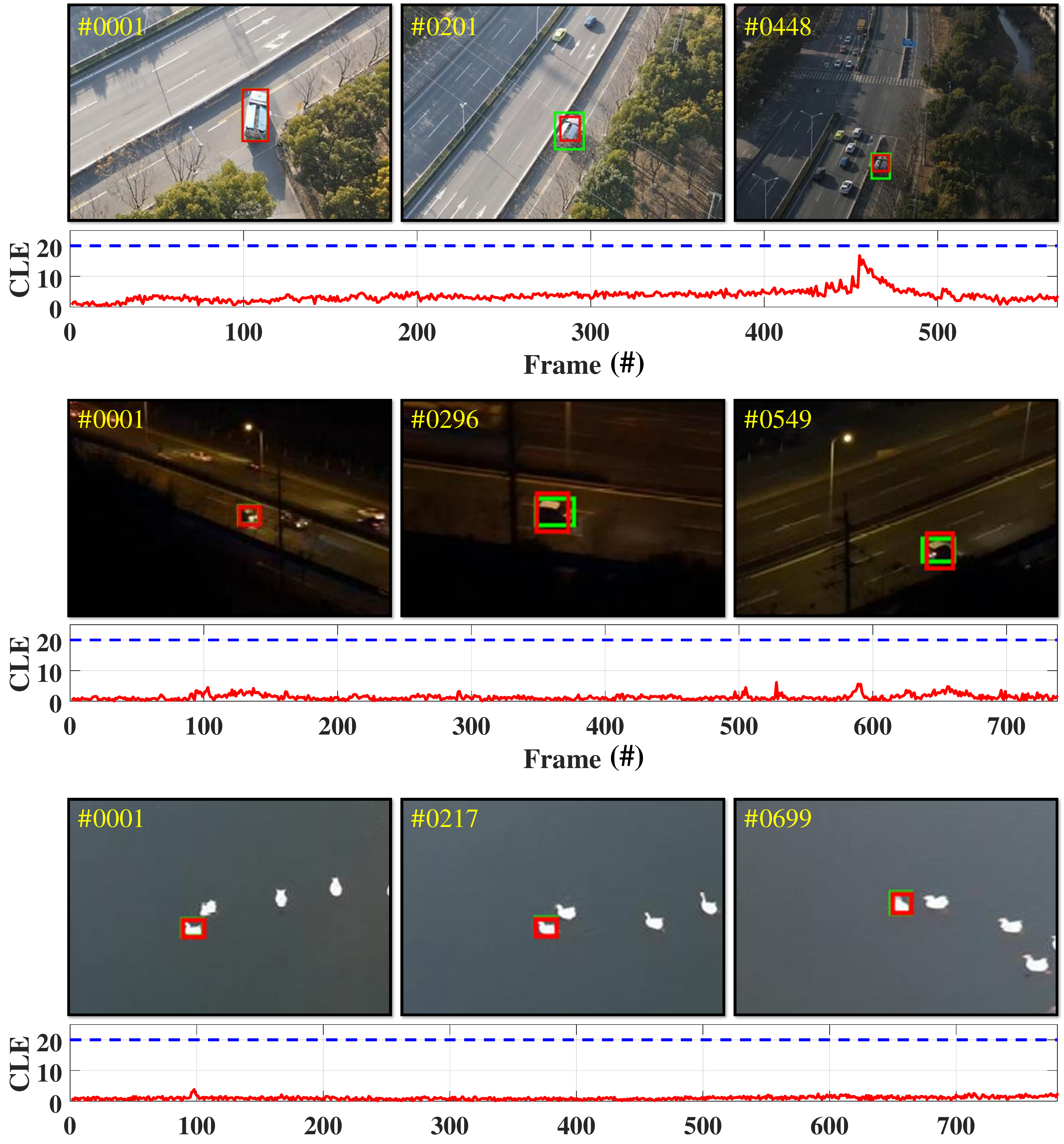}
		
		\caption{Visualization of real-world tests on the embedded platform. The tracking results and ground truth are marked with \NoOne{red} and \NoTwo{green} boxes. The CLE score below the \NoThree{blue} dotted line is considered as the success tracking result in the real-world tests.} 
		\vspace{-16pt}
		\label{fig:v4r}
	\end{figure}

	In this section, HiFT is further implemented on a typical UAV platform including an embedded onboard processor, \textit{i.e.}, NVIDIA AGX Xavier, to testify its practicability in real-world applications. Figure~\ref{fig:v4r} presents three tests in the wild, including day and night scenes.
	The main challenges in the tests are partial occlusion, viewpoint change (the first row), low-resolution, camera motion (the second row), small object, and similar object around (the third row). Attributing to the effective feature transformer, HiFT maintains satisfying tracking robustness in various challenging scenarios. Moreover, our tracker remains at an average speed of \textbf{31.2} FPS during the tests without using TensorRT. Therefore, the real-world tests onboard the embedded system directly validate the superior performance and efficiency of HiFT under various UAV-specific challenges. 
	
	\section{Conclusion}\label{sec:CONCLUSIONS}
	In this work, a novel hierarchical feature transformer for efficient aerial tracking is proposed for streamlining the process of exploiting the global contextual information and multi-level features. By virtue of both low-resolution semantics information and high-resolution spatial details, the transformed feature can achieve promising performance in discriminating the object from clutters via a lightweight structure.  Meanwhile, attributing to the modulation layer and the new classification label, the effectiveness of the feature transformer can reach its full potential. Comprehensive experiments have validated that HiFT can achieve an excellent precision-speed trade-off and can be utilized in real-world aerial tracking scenarios. Moreover, even compared to the trackers with deeper backbones, HiFT can achieve comparable performance. We are convinced that our work can advance the development of aerial tracking and promote the real-world applications of visual tracking.
	
	\textbf{Acknowledgment:} This work is supported by the National Natural Science Foundation of China (No. 61806148) and the Natural Science Foundation of Shanghai (No. 20ZR1460100). We thank the anonymous reviewers for their efforts to help us improve our work.
	
	{\small
		\bibliographystyle{ieee_fullname}
		\bibliography{egbib}

\begin{thebibliography}{10}\itemsep=-1pt

\bibitem{vaswani2017attention}
V. {Ashish}, S. {Noam}, P. {Niki}, U. {Jakob}, J. {Llion}, N.~G. {Aidan}, K.
  {Lukasz}, and P. {Illia}.
\newblock {Attention Is All You Need}.
\newblock In {\em Advances in neural information processing systems (NIPS)},
  pages 6000--6010, 2017.

\bibitem{bertinettofully}
L. Bertinetto, J. Valmadre, J.~F. Henriques, A. Vedaldi, and P.~H. Torr.
\newblock {Fully-Convolutional Siamese Networks for Object Tracking}.
\newblock In {\em Proceedings of the European Conference on Computer Vision
  (ECCV)}, pages 850--865, 2016.

\bibitem{9010649}
G. {Bhat}, M. {Danelljan}, L. {Van Gool}, and R. {Timofte}.
\newblock Learning discriminative model prediction for tracking.
\newblock In {\em Proceedings of the IEEE International Conference on Computer
  Vision (ICCV)}, pages 6181--6190, 2019.

\bibitem{5539960}
D.~S. {Bolme}, J.~R. {Beveridge}, B.~A. {Draper}, and Y.~M. {Lui}.
\newblock {Visual Object Tracking Using Adaptive Correlation Filters}.
\newblock In {\em Proceedings of the IEEE Conference on Computer Vision and
  Pattern Recognition (CVPR)}, pages 2544--2550, 2010.

\bibitem{8968163}
R. {Bonatti}, C. {Ho}, W. {Wang}, S. {Choudhury}, and S. {Scherer}.
\newblock {Towards a Robust Aerial Cinematography Platform: Localizing and
  Tracking Moving Targets in Unstructured Environments}.
\newblock In {\em Proceedings of the IEEE/RSJ International Conference on
  Intelligent Robots and Systems (IROS)}, pages 229--236, 2019.

\bibitem{cao2021siamapn++}
Z. {Cao}, C. {Fu}, J. {Ye}, B. {Li}, and Y. {Li}.
\newblock {SiamAPN++: Siamese Attentional Aggregation Network for Real-Time UAV
  Tracking}.
\newblock In {\em Proceedings of the IEEE/RSJ International Conference on
  Intelligent Robots and Systems (IROS)}, pages 1--7, 2021.

\bibitem{carion2020end}
N. Carion, F. Massa, G. Synnaeve, N. Usunier, A. Kirillov, and S. Zagoruyko.
\newblock {End-to-End Object Detection with Transformers}.
\newblock In {\em Proceedings of the European Conference on Computer Vision
  (ECCV)}, pages 213--229, 2020.

\bibitem{chen2020siamese}
Z. Chen, B. Zhong, G. Li, S. Zhang, and R. Ji.
\newblock {Siamese Box Adaptive Network for Visual Tracking}.
\newblock In {\em Proceedings of the IEEE Conference on Computer Vision and
  Pattern Recognition (CVPR)}, pages 6668--6677, 2020.

\bibitem{8100216}
M. {Danelljan}, G. {Bhat}, F.~S. {Khan}, and M. {Felsberg}.
\newblock {ECO: Efficient Convolution Operators for Tracking}.
\newblock In {\em Proceedings of the IEEE Conference on Computer Vision and
  Pattern Recognition (CVPR)}, pages 6931--6939, 2017.

\bibitem{8953466}
M. {Danelljan}, G. {Bhat}, F.~S. {Khan}, and M. {Felsberg}.
\newblock {ATOM: Accurate Tracking by Overlap Maximization}.
\newblock In {\em Proceedings of the IEEE Conference on Computer Vision and
  Pattern Recognition (CVPR)}, pages 4655--4664, 2019.

\bibitem{7410847}
M. {Danelljan}, G. {Häger}, F.~S. {Khan}, and M. {Felsberg}.
\newblock {Learning Spatially Regularized Correlation Filters for Visual
  Tracking}.
\newblock In {\em Proceedings of the IEEE International Conference on Computer
  Vision (ICCV)}, pages 4310--4318, 2015.

\bibitem{7569092}
M. {Danelljan}, G. {Häger}, F.~S. {Khan}, and M. {Felsberg}.
\newblock {Discriminative Scale Space Tracking}.
\newblock {\em IEEE Transactions on Pattern Analysis and Machine Intelligence},
  39(8):1561--1575, 2017.

\bibitem{danelljan2016beyond}
M. Danelljan, A. Robinson, F.~S. Khan, and M. Felsberg.
\newblock {Beyond Correlation Filters: Learning Continuous Convolution
  Operators for Visual Tracking}.
\newblock In {\em Proceedings of the European Conference on Computer Vision
  (ECCV)}, pages 472--488, 2016.

\bibitem{9157124}
M. {Danelljan}, L. {Van Gool}, and R. {Timofte}.
\newblock {Probabilistic Regression for Visual Tracking}.
\newblock In {\em Proceedings of the IEEE Conference on Computer Vision and
  Pattern Recognition (CVPR)}, pages 7181--7190, 2020.

\bibitem{real2017youtube}
E.{Real}, J.{Shlens}, S.{Mazzocchi}, X.{Pan}, and V.{Vanhoucke}.
\newblock {YouTube-BoundingBoxes: A Large High-Precision Human-Annotated Data
  Set for Object Detection in Video}.
\newblock In {\em Proceedings of the IEEE Conference on Computer Vision and
  Pattern Recognition (CVPR)}, pages 7464--7473, 2017.

\bibitem{8954057}
H. {Fan} and H. {Ling}.
\newblock {Siamese Cascaded Region Proposal Networks for Real-Time Visual
  Tracking}.
\newblock In {\em Proceedings of the IEEE Conference on Computer Vision and
  Pattern Recognition (CVPR)}, pages 7944--7953, 2019.

\bibitem{9477413}
C. {Fu}, Z. {Cao}, Y. {Li}, J. {Ye}, and C. {Feng}.
\newblock {Onboard Real-Time Aerial Tracking With Efficient Siamese Anchor
  Proposal Network}.
\newblock {\em IEEE Transactions on Geoscience and Remote Sensing}, pages
  1--13, 2021.

\bibitem{fu2020siamese}
C. {Fu}, Z. {Cao}, Y. {Li}, J. {Ye}, and C. {Feng}.
\newblock {Siamese Anchor Proposal Network for High-Speed Aerial Tracking}.
\newblock In {\em Proceedings of the IEEE International Conference on Robotics
  and Automation (ICRA)}, pages 1--7, 2021.

\bibitem{6907659}
C. {Fu}, A. {Carrio}, M.~A. {Olivares-Mendez}, R. {Suarez-Fernandez}, and P.
  {Campoy}.
\newblock {Robust real-time vision-based aircraft tracking from Unmanned Aerial
  Vehicles}.
\newblock In {\em Proceedings of the IEEE International Conference on Robotics
  and Automation (ICRA)}, pages 5441--5446, 2014.

\bibitem{fu2020correlation}
Changhong Fu, Bowen Li, Fangqiang Ding, Fuling Lin, and Geng Lu.
\newblock {Correlation Filter for UAV-Based Aerial Tracking: A Review and
  Experimental Evaluation}.
\newblock {\em IEEE Geoscience and Remote Sensing Magazine}, pages 1--28, 2020.

\bibitem{kiani2017learning}
H.~K. {Galoogahi}, A. {Fagg}, and S. {Lucey}.
\newblock {Learning Background-Aware Correlation Filters for Visual Tracking}.
\newblock In {\em Proceedings of the IEEE International Conference on Computer
  Vision (ICCV)}, pages 1144--1152, 2017.

\bibitem{Guo_2021_CVPR}
D. {Guo}, Y. {Shao}, Y. {Cui}, Z. {Wang}, L. {Zhang}, and C. {Shen}.
\newblock {Graph Attention Tracking}.
\newblock In {\em Proceedings of the IEEE Conference on Computer Vision and
  Pattern Recognition (CVPR)}, pages 1--10, 2021.

\bibitem{9157720}
D. {Guo}, J. {Wang}, Y. {Cui}, Z. {Wang}, and S. {Chen}.
\newblock {SiamCAR: Siamese Fully Convolutional Classification and Regression
  for Visual Tracking}.
\newblock In {\em Proceedings of the IEEE Conference on Computer Vision and
  Pattern Recognition (CVPR)}, pages 6268--6276, 2020.

\bibitem{guo2017learning}
Q. {Guo}, W. {Feng}, C. {Zhou}, R. {Huang}, L. {Wan}, and S. {Wang}.
\newblock {Learning Dynamic Siamese Network for Visual Object Tracking}.
\newblock In {\em Proceedings of the IEEE International Conference on Computer
  Vision (ICCV)}, pages 1781--1789, 2017.

\bibitem{He2016res}
K. {He}, X. {Zhang}, S. {Ren}, and J. {Sun}.
\newblock {Deep Residual Learning for Image Recognition}.
\newblock In {\em Proceedings of the IEEE Conference on Computer Vision and
  Pattern Recognition (CVPR)}, pages 770--778, 2016.

\bibitem{6870486}
J.~F. {Henriques}, R. {Caseiro}, P. {Martins}, and J. {Batista}.
\newblock {High-Speed Tracking with Kernelized Correlation Filters}.
\newblock {\em IEEE Transactions on Pattern Analysis and Machine Intelligence},
  37(3):583--596, 2015.

\bibitem{huang2020hand}
L. {Huang}, J. {Tan}, J. {Liu}, and J. {Yuan}.
\newblock {Hand-Transformer: Non-Autoregressive Structured Modeling for 3D Hand
  Pose Estimation}.
\newblock In {\em Proceedings of the European Conference on Computer Vision
  (ECCV)}, pages 17--33, 2020.

\bibitem{huang2019got}
L. {Huang}, X. {Zhao}, and K. {Huang}.
\newblock {GOT-10k: A Large High-Diversity Benchmark for Generic Object
  Tracking in the Wild}.
\newblock {\em IEEE Transactions on Pattern Analysis and Machine Intelligence},
  pages 1--17, 2019.

\bibitem{huang2019learning}
Z. {Huang}, C. {Fu}, Y. {Li}, F. {Lin}, and P. {Lu}.
\newblock {Learning Aberrance Repressed Correlation Filters for Real-time UAV
  Tracking}.
\newblock In {\em Proceedings of the IEEE International Conference on Computer
  Vision (ICCV)}, pages 2891--2900, 2019.

\bibitem{krizhevsky2012imagenet}
A. {Krizhevsky}, I. {Sutskever}, and G.~E. {Hinton}.
\newblock {Imagenet Classification with Deep Convolutional Neural Networks}.
\newblock In {\em Advances in Neural Information Processing Systems (NeurIPS)},
  pages 1097--1105, 2012.

\bibitem{8954116}
B. {Li}, W. {Wu}, Q. {Wang}, F. {Zhang}, J. {Xing}, and J. {Yan}.
\newblock {SiamRPN++: Evolution of Siamese Visual Tracking With Very Deep
  Networks}.
\newblock In {\em Proceedings of the IEEE Conference on Computer Vision and
  Pattern Recognition (CVPR)}, pages 4277--4286, 2019.

\bibitem{8579033}
B. {Li}, J. {Yan}, W. {Wu}, Z. {Zhu}, and X. {Hu}.
\newblock {High Performance Visual Tracking with Siamese Region Proposal
  Network}.
\newblock In {\em Proceedings of the IEEE Conference on Computer Vision and
  Pattern Recognition (CVPR)}, pages 8971--8980, 2018.

\bibitem{8578613}
F. {Li}, C. {Tian}, W. {Zuo}, L. {Zhang}, and M. {Yang}.
\newblock {Learning Spatial-Temporal Regularized Correlation Filters for Visual
  Tracking}.
\newblock In {\em Proceedings of the IEEE Conference on Computer Vision and
  Pattern Recognition (CVPR)}, pages 4904--4913, 2018.

\bibitem{li2017visual}
S. {Li} and D. {Yeung}.
\newblock {Visual Object Tracking for Unmanned Aerial Vehicles: A Benchmark and
  New Motion Models}.
\newblock In {\em Proceedings of the AAAI Conference on Artificial Intelligence
  (AAAI)}, pages 1--7, 2017.

\bibitem{Xin2019CVPR}
X. {Li}, C. {Ma}, B. {Wu}, Z. {He}, and M. {Yang}.
\newblock {Target-Aware Deep Tracking}.
\newblock In {\em Proceedings of the IEEE Conference on Computer Vision and
  Pattern Recognition (CVPR)}, pages 1369--1378, 2019.

\bibitem{Li_2020_CVPR}
Y. {Li}, C. {Fu}, F. {Ding}, Z. {Huang}, and G. {Lu}.
\newblock {AutoTrack: Towards High-Performance Visual Tracking for UAV With
  Automatic Spatio-Temporal Regularization}.
\newblock In {\em Proceedings of the IEEE Conference on Computer Vision and
  Pattern Recognition (CVPR)}, pages 11920--11929, 2020.

\bibitem{lin2014microsoft}
T. {Lin}, M. {Maire}, S. {Belongie}, J. {Hays}, P. {Perona}, D. {Ramanan}, P.
  Doll{\'a}r, and C.~L. {Zitnick}.
\newblock {Microsoft coco: Common objects in context}.
\newblock In {\em Proceedings of the European conference on computer vision
  (ECCV)}, pages 740--755, 2014.

\bibitem{meinhardt2021trackformer}
T. {Meinhardt}, A. {Kirillov}, L. {Laura}, and C. {Feichtenhofer}.
\newblock {TrackFormer: Multi-Object Tracking with Transformers}.
\newblock {\em arXiv preprint arXiv:2101.02702}, 2021.

\bibitem{Mueller2016ECCV}
M. {Mueller}, N. {Smith}, and B. {Ghanem}.
\newblock {A Benchmark and Simulator for UAV Tracking}.
\newblock In {\em Proceedings of the European Conference on Computer Vision
  (ECCV)}, pages 445--461, 2016.

\bibitem{girshick2015fast}
S. {Ren}, K. {He}, R. {Girshick}, and J. {Sun}.
\newblock {Faster R-CNN: Towards Real-Time Object Detection with Region
  Proposal Networks}.
\newblock In {\em {Advances in Neural Information Processing Systems
  (NeurIPS)}}, pages 91--99, 2015.

\bibitem{russakovsky2015imagenet}
O. {Russakovsky}, J. {Deng}, H. {Su}, J. {Krause}, et~al.
\newblock {Imagenet Large Scale Visual Recognition Challenge}.
\newblock {\em International Journal of Computer Vision}, 115(3):211--252,
  2015.

\bibitem{sandler2018mobilenetv2}
M. {Sandler}, A. {Howard}, M. {Zhu}, A. {Zhmoginov}, and L. {Chen}.
\newblock {Mobilenetv2: Inverted Residuals and Linear Bottlenecks}.
\newblock In {\em Proceedings of the IEEE Conference on Computer Vision and
  Pattern Recognition (CVPR)}, pages 4510--4520, 2018.

\bibitem{szegedy2015going}
Christian Szegedy, Wei Liu, Yangqing Jia, Pierre Sermanet, Scott Reed, Dragomir
  Anguelov, Dumitru Erhan, Vincent Vanhoucke, and Andrew Rabinovich.
\newblock {Going Deeper with Convolutions}.
\newblock In {\em Proceedings of the IEEE Conference on Computer Vision and
  Pattern Recognition (CVPR)}, pages 1--9, 2015.

\bibitem{Wang_2019_Unsupervised}
N. {Wang}, Y. {Song}, C. {Ma}, W. {Zhou}, W. {Liu}, and H. {Li}.
\newblock {Unsupervised Deep Tracking}.
\newblock In {\em Proceedings of the IEEE Conference on Computer Vision and
  Pattern Recognition (CVPR)}, pages 1308--1317, 2019.

\bibitem{8578607}
N. {Wang}, W. {Zhou}, Q. {Tian}, R. {Hong}, M. {Wang}, and H. {Li}.
\newblock {Multi-cue Correlation Filters for Robust Visual Tracking}.
\newblock In {\em Proceedings of the IEEE Conference on Computer Vision and
  Pattern Recognition}, pages 4844--4853, 2018.

\bibitem{8953931}
Q. {Wang}, L. {Zhang}, L. {Bertinetto}, W. {Hu}, and P.~H.~S. {Torr}.
\newblock Fast online object tracking and segmentation: A unifying approach.
\newblock In {\em Proceedings of the IEEE Conference on Computer Vision and
  Pattern Recognition (CVPR)}, pages 1328--1338, 2019.

\bibitem{yang2020learning}
F. {Yang}, H. {Yang}, J. {Fu}, H. {Lu}, and B. {Guo}.
\newblock {Learning Texture Transformer Network for Image Super-Resolution}.
\newblock In {\em Proceedings of the IEEE Conference on Computer Vision and
  Pattern Recognition (CVPR)}, pages 5791--5800, 2020.

\bibitem{9457090}
J. {Ye}, C. {Fu}, F. {Lin}, F. {Ding}, S. {An}, and G. {Lu}.
\newblock {Multi-Regularized Correlation Filter for UAV Tracking and
  Self-Localization}.
\newblock {\em IEEE Transactions on Industrial Electronics}, pages 1--10, 2021.

\bibitem{yu2015multi}
F. {Yu} and V. {Koltun}.
\newblock {Multi-Scale Context Aggregation by Dilated Convolutions}.
\newblock In {\em Proceedings of the International Conference on Learning
  Representations (ICLR)}, pages 1--9, 2016.

\bibitem{zhang2017robust}
L. {Zhang} and P.~N. Suganthan.
\newblock {Robust Visual Tracking via Co-Trained Kernelized Correlation
  Filters}.
\newblock {\em Pattern Recognition}, 69:82--93, 2017.

\bibitem{8579009}
L. {Zhou}, Y. {Zhou}, J.~J. {Corso}, R. {Socher}, and C. {Xiong}.
\newblock {End-to-End Dense Video Captioning with Masked Transformer}.
\newblock In {\em Proceedings of the IEEE Conference on Computer Vision and
  Pattern Recognition (CVPR)}, pages 8739--8748, 2018.

\bibitem{zhu2020deformable}
X. Zhu, W. Su, L. Lu, B. Li, X. Wang, and J. Dai.
\newblock {Deformable DETR: Deformable Transformers for End-to-End Object
  Detection}.
\newblock {\em arXiv preprint arXiv:2010.04159}, 2020.

\bibitem{zhu2018distractor}
Z. Zhu, Q. Wang, B. Li, W. Wu, J. Yan, and W. Hu.
\newblock {Distractor-Aware Siamese Networks for Visual Object Tracking}.
\newblock In {\em Proceedings of the European Conference on Computer Vision
  (ECCV)}, pages 101--117, 2018.

\end{thebibliography}
	}
	
\end{document}